\documentclass[11pt]{article}

\usepackage[preprint]{acl}

\usepackage{times}
\usepackage{latexsym}

\usepackage[T1]{fontenc}

\usepackage[utf8]{inputenc}

\usepackage{microtype}

\usepackage{inconsolata}

\usepackage{graphicx}

\usepackage{url}            
\usepackage{booktabs}       
\usepackage{amsfonts}       
\usepackage{nicefrac}       
\usepackage{microtype}      
\usepackage{listings}
\usepackage{caption}
\usepackage{graphicx}
\usepackage{makecell}
\usepackage{subcaption}
\usepackage{microtype}
\usepackage{multirow}
\usepackage{bm}
\usepackage{tabularx}
\usepackage{ltablex}
\usepackage{longtable}
\usepackage{colortbl}
\usepackage{supertabular}
\usepackage{xltabular}
\usepackage{adjustbox}
\usepackage{lineno}
\usepackage{array}
\usepackage{booktabs}
\usepackage{amsmath}
\usepackage{amsfonts}
\usepackage{amssymb}
\usepackage{makecell}
\usepackage{paralist}
\usepackage[most]{tcolorbox}
\usepackage{algorithm}
\usepackage{algpseudocode}
\usepackage{multirow}
\usepackage{etoolbox}
\usepackage{mdframed}

\newmdenv[
    linecolor=blue,
    linewidth=2pt,
    topline=true,
    bottomline=true,
    leftline=true,
    rightline=true,
    backgroundcolor=blue!10
]{bluebox}

\usepackage{amsmath}
\usepackage{amssymb}
\usepackage{mathtools}
\usepackage{amsthm}
\usepackage{pifont}
\usepackage[dvipsnames]{xcolor}

\usepackage[capitalize,noabbrev]{cleveref}

\newcommand{\vh}{\mathbf{h}}          
\newcommand{\vx}{\mathbf{x}}          
\newcommand{\vd}{\mathbf{d}}          
\newcommand{\vb}{\mathbf{b}}          

\newcommand{\mR}{\mathbf{R}}          
\newcommand{\mI}{\mathbf{I}}          

\newtheorem{proposition}{Proposition}
\allowdisplaybreaks

\title{Selective Steering: Norm-Preserving Control Through Discriminative Layer Selection}

\author{Quy-Anh Dang$^{1,2}$, Chris Ngo$^{2}$ \\
$^1$VNU University of Science, Vietnam \\
$^2$Knovel Engineering Lab, Singapore \\
\texttt{\{quyanh.dang, chris.ngo\}@knoveleng.com}
\vspace{1em}  
\\
\textbf{Project:} \href{https://knoveleng.github.io/steering/}{https://knoveleng.github.io/steering/} \\
}

\begin{document}
\maketitle
\begin{abstract}
Despite significant progress in alignment, large language models (LLMs) remain vulnerable to adversarial attacks that elicit harmful behaviors. Activation steering techniques offer a promising inference-time intervention approach, but existing methods suffer from critical limitations: activation addition requires careful coefficient tuning and is sensitive to layer-specific norm variations, while directional ablation provides only binary control. Recent work on Angular Steering introduces continuous control via rotation in a 2D subspace, but its practical implementation violates norm preservation, causing distribution shift and generation collapse, particularly in models below 7B parameters. We propose \textbf{Selective Steering}\footnote{\textbf{Code:} \href{https://github.com/knoveleng/steering}{https://github.com/knoveleng/steering}}, which addresses these limitations through two key innovations: (1) a mathematically rigorous norm-preserving rotation formulation that maintains activation distribution integrity, and (2) discriminative layer selection that applies steering only where feature representations exhibit opposite-signed class alignment. Experiments across nine models demonstrate that Selective Steering achieves 5.5$\times$ higher attack success rates than prior methods while maintaining zero perplexity violations and approximately 100\% capability retention on standard benchmarks. Our approach provides a principled, efficient framework for controllable and stable LLM behavior modification.

\end{abstract}


\section{Introduction}
\label{sec:introduction}

Large Language Models (LLMs) have demonstrated remarkable capabilities, yet ensuring their safe deployment remains critical. Despite extensive alignment efforts through RLHF~\citep{ouyang2022training} and constitutional AI~\citep{bai2022constitutional}, models remain vulnerable to jailbreaks~\citep{zou2023universal} and harmful behaviors~\citep{perez-etal-2022-red}. Traditional alignment requires expensive retraining and often degrades performance on benign tasks~\citep{casper2023open, tan2025equilibraterlhfbalancinghelpfulnesssafety}.

\begin{figure}[t!]
    \centering
    \includegraphics[width=\linewidth]{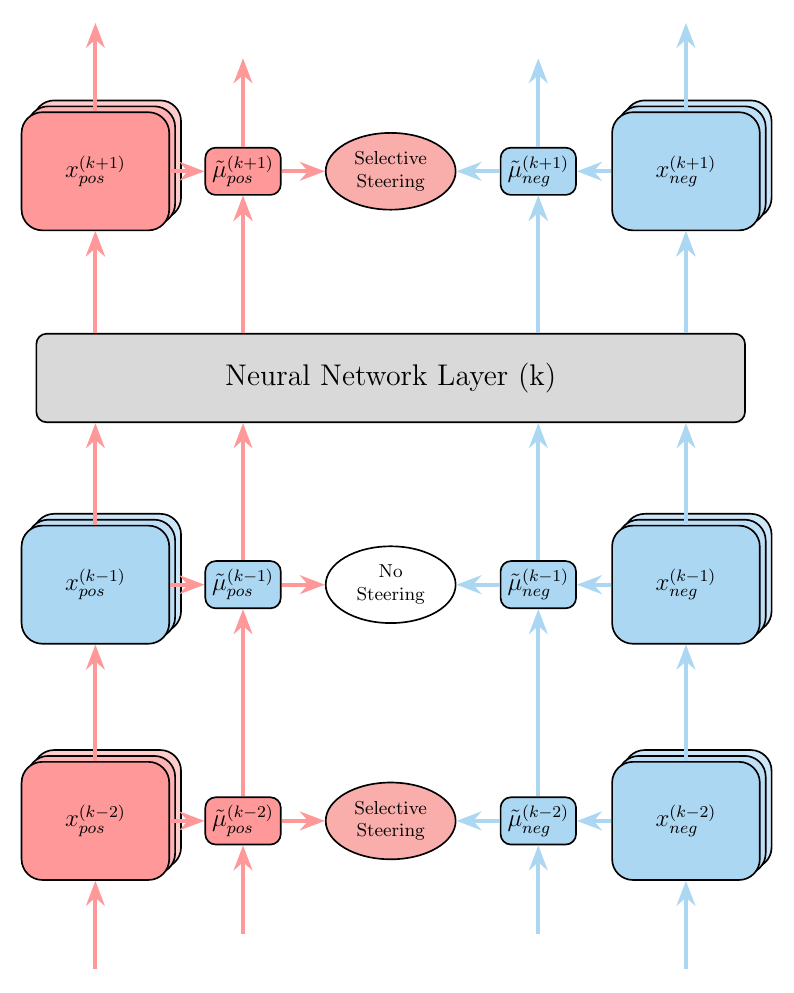}
    \caption{%
        \textbf{Selective Steering pipeline.} 
        At each layer $k$, we compute projections of positive (\textcolor{red!40}{red}) and negative (\textcolor{CornflowerBlue!50}{blue}) class means onto the selected  feature direction (\textcolor{red!40}{red}/\textcolor{CornflowerBlue!50}{blue} boxes). 
        Steering is applied only at layers where projections have opposite signs (layers $k-2$ and $k+1$), using norm-preserving rotation. 
        Layers with same-sign projections (layer $k-1$) remain unchanged.
    }
    \label{fig:diagram}
\end{figure}

\noindent \textbf{Activation steering} - modifying internal representations at inference time - offers an alternative~\citep{turner2024steeringlanguagemodelsactivation, zou2023transparency}. However, existing methods face critical limitations: \textbf{Activation Addition} requires careful coefficient tuning and is sensitive to layer-specific norms~\citep{templeton2024scaling}, while \textbf{Directional Ablation} removes features entirely, precluding fine-grained control~\citep{arditi2024refusal}. Recent \textbf{Angular Steering}~\citep{vu2025angular} reformulates steering as geometric rotation in a 2D subspace, but suffers from \emph{generation collapse on small models (<7B)} and \emph{poor controllability on strongly aligned models} (Qwen, Gemma).

\paragraph{Our Approach.}
We hypothesize these failures stem from \textbf{uniform steering across all layers}, ignoring heterogeneous layer roles. Through systematic analysis, we identify: (1) non-uniform activation norm growth across depth; (2) progressive emergence of opposite-signed discriminability in middle-to-late layers; and (3) layer-specific vulnerability to steering.

We propose \textbf{Selective Steering (SS)}, which applies norm-preserving rotation \emph{only to layers where contrastive classes exhibit opposite-signed projections}: $\boldsymbol{\tilde{\mu}}^{(k)}_{\text{pos}} \cdot \boldsymbol{\tilde{\mu}}^{(k)}_{\text{neg}}$. This discriminative criterion identifies \emph{steerable layers} where features are meaningfully represented, achieving: (1) maintained coherence by avoiding non-discriminative layers; (2) enhanced controllability by concentrating effort where separation emerges; and (3) preserved general capabilities.

\paragraph{Contributions.} Our contributions are threefold:
\begin{compactenum}
    \item We provide the first systematic analysis of layer-wise activation geometry in the context of steering, identifying non-uniform norm growth and progressive discriminability emergence as key phenomena governing steering effectiveness.
    
    \item We propose Selective Steering, a principled method that combines norm-preserving rotation with discriminative layer selection. We prove that SS guarantees activation norm preservation (Proposition~\ref{prop:norm_preservation}) while standard Angular Steering violates this property (Proposition~\ref{prop:angular_violation}).
    
    \item Through comprehensive experiments on 8 models across 3 families (Llama, Qwen, Gemma), we demonstrate that SS simultaneously achieves: (1) zero perplexity threshold violations across all models and angles; (2) up to 5.5× improvement in attack success rate on challenging models; and (3) preservation of general capabilities, substantially outperforming existing methods.
\end{compactenum}


\section{Background}
\label{sec:background}

\subsection{Transformer Architecture}
Decoder-only transformers process an input token sequence $\mathbf{t} = (t_1, \dots, t_n)$ by first converting tokens to initial embeddings, $\vh^{(1)}_i = \text{Embed}(t_i)$, where $\vh$ denotes a vector in activation space. These activations are then iteratively refined through $L$ layers via a residual stream architecture. Within each layer $\ell$, the residual stream activation $\vh^{(\ell)}_i$ for token $t_i$ is updated by incorporating information from a self-attention mechanism and a multi-layer perceptron (MLP) block, typically with normalization applied before these components:
\begin{align}
\vh^{(\ell)}_{i, \text{post-attn}} &= \vh^{(\ell)}_i + \text{Attn}^{(\ell)}(\text{Norm}(\vh^{(\ell)}_{1:i})) \nonumber \\
\vh^{(\ell+1)}_i &= \vh^{(\ell)}_{i, \text{post-attn}} + \text{MLP}^{(\ell)}(\text{Norm}(\vh^{(\ell)}_{i, \text{post-attn}}))
\end{align}
This layered processing constructs increasingly sophisticated representations, where $\vh \in \mathbb{R}^{d_\text{model}}$. Finally, output activations from the last layer, $\vh^{(L+1)}_i$, are projected to vocabulary logits via $\text{logits}_i = \text{Unembed}(\vh^{(L+1)}_i)$, which are then normalized using softmax to produce probability distributions $\mathbf{y}_i$ for next-token prediction.

\subsection{Activation Steering}
Activation steering modifies internal model representations at inference time to induce or suppress specific behaviors without requiring retraining~\citep{turner2024steeringlanguagemodelsactivation, arditi2024refusal}. Features are hypothesized to be represented by orthogonal directions in activation space~\citep{elhage2022toymodelssuperposition}, enabling targeted interventions through geometric transformations. Existing methods include vector addition~\citep{turner2024steeringlanguagemodelsactivation}, orthogonal projection~\citep{arditi2024refusal}, and geometric rotation~\citep{vu2025angular}. A comprehensive comparison of these approaches is provided in Appendix ~\ref{sec:related}.

\paragraph{Angular Steering Framework.}
We build upon Angular Steering~\citep{vu2025angular}, which reformulates activation editing as rotation within a 2D subspace. Given an orthonormal basis $\{\vb_1, \vb_2\}$ spanning the steering plane $P$, rotation to target angle $\theta$ is implemented as:
\begin{align}
\label{eq:angular_steering_transform}
\vh_{\text{steered}, \theta} &= \vh - \text{proj}_P(\vh) \notag \\
&+ \|\text{proj}_P(\vh)\| \cdot [\vb_1 \; \vb_2] \, \mR_{\theta} \, [1 \; 0]^\top,
\end{align}
where $\text{proj}_P(\vh) = (\vb_1\vb_1^\top + \vb_2\vb_2^\top)\vh$ denotes the projection of $\vh$ onto the steering plane, and $\mR_\theta$ is the standard 2D rotation matrix:
\begin{align}
\mR_\theta = \begin{bmatrix} \cos(\theta) & -\sin(\theta) \\ \sin(\theta) & \cos(\theta) \end{bmatrix}.
\end{align}
This formulation provides continuous control over behavioral intensity through the rotation angle $\theta \in [0°, 360°)$.

\subsection{Feature Direction Extraction}
\label{subsec:feature_extraction}
The most established method for constructing steering vectors is the \textit{difference-in-means} approach~\citep{Belrose2023DiffInMeans}. Given contrastive prompt sets - a \textit{negative} set $\mathcal{D}^{(\text{train})}_{\text{neg}}$ where a target feature is absent and a \textit{positive} set $\mathcal{D}^{(\text{train})}_{\text{pos}}$ where the feature is present - the steering vector at layer $k$ is computed as:
\begin{align}
    \vd^{(k)} = \boldsymbol{\mu}^{(k)}_{\text{pos}} - \boldsymbol{\mu}^{(k)}_{\text{neg}},
\end{align}
where the class-conditional mean vectors are:
\begin{align}
    \boldsymbol{\mu}^{(k)}_{\text{pos}} = \frac{1}{|\mathcal{D}^{(\text{train})}_{\text{pos}}|} 
    \sum_{p \in \mathcal{D}^{(\text{train})}_{\text{pos}}} \vx^{(k)}(p), \notag \\
    \boldsymbol{\mu}^{(k)}_{\text{neg}} = \frac{1}{|\mathcal{D}^{(\text{train})}_{\text{neg}}|} 
    \sum_{p \in \mathcal{D}^{(\text{train})}_{\text{neg}}} \vx^{(k)}(p). \label{eq:diff-in-mean}
\end{align}
Here, $\vx^{(k)}(p)$ denotes the activation vector at layer $k$ for prompt $p$. This difference vector $\vd^{(k)}$ points in the direction that maximally separates the two classes in activation space. We normalize it to obtain the unit steering direction: $\hat{\vd}^{(k)} = \vd^{(k)} / \|\vd^{(k)}\|$.

\section{Methodology}
\label{sec:methodology}

\subsection{Limitations of Angular Steering}

While Angular Steering~\citep{vu2025angular} introduces continuous control through rotation in a 2D subspace, its practical implementation suffers from a critical flaw: \textbf{norm distortion}. Although the theoretical rotation matrix is mathematically sound, the efficient implementation (Equation~\ref{eq:angular_steering_transform}) fails to preserve norms.

\begin{proposition}[Norm Violation in Angular Steering]
\label{prop:angular_violation}
The Angular Steering implementation (Equation~\ref{eq:angular_steering_transform}) does not preserve activation norms for general rotation angles $\theta$.
\end{proposition}

We provide a constructive proof in Appendix~\ref{app:proof_angular_violation}, demonstrating that even at $\theta = 0°$ (the identity transformation), norm preservation fails unless the activation's projection onto the steering plane lies exactly along $\vb_1$ with non-negative coefficient. This violation propagates through Adaptive Angular Steering, which inherits the same transformation.

\paragraph{Consequences.}
Norm distortion becomes particularly problematic in modern LLMs employing normalization layers (LayerNorm~\citep{ba2016layer}, RMSNorm~\citep{zhang2019root}), leading to: (1) distribution shift as activations fall outside expected norms; (2) accumulation of distortions across layers; (3) unpredictable steering strength varying by layer and prompt.

\subsection{Empirical Observations: Layer-Wise Heterogeneity}

\begin{figure*}[t]
  \centering
  \subfloat[Activation norms across layers\label{fig:activation_norms}]{
    \includegraphics[width=0.48\textwidth]{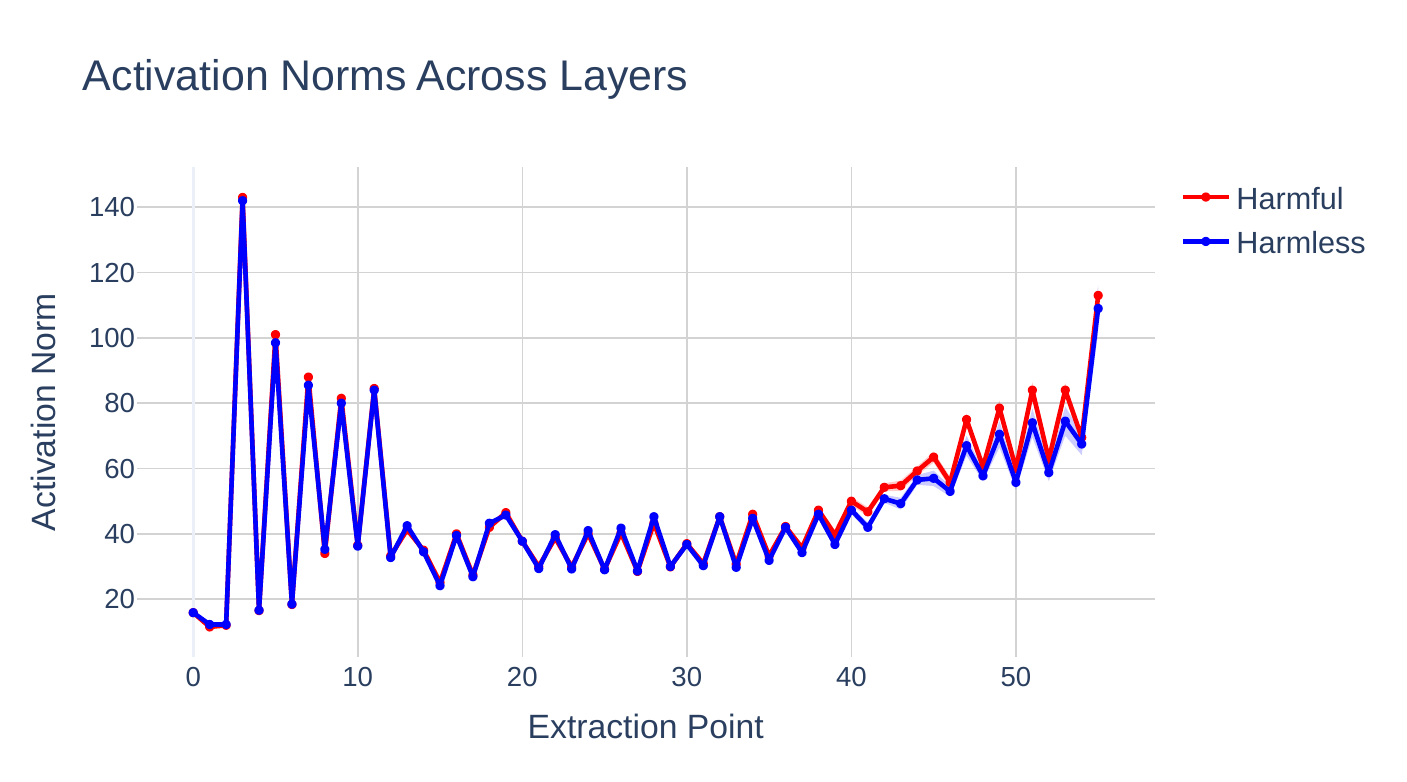}
  }\hfill
  \subfloat[Alignment with selected feature direction\label{fig:projections_local}]{
    \includegraphics[width=0.48\textwidth]{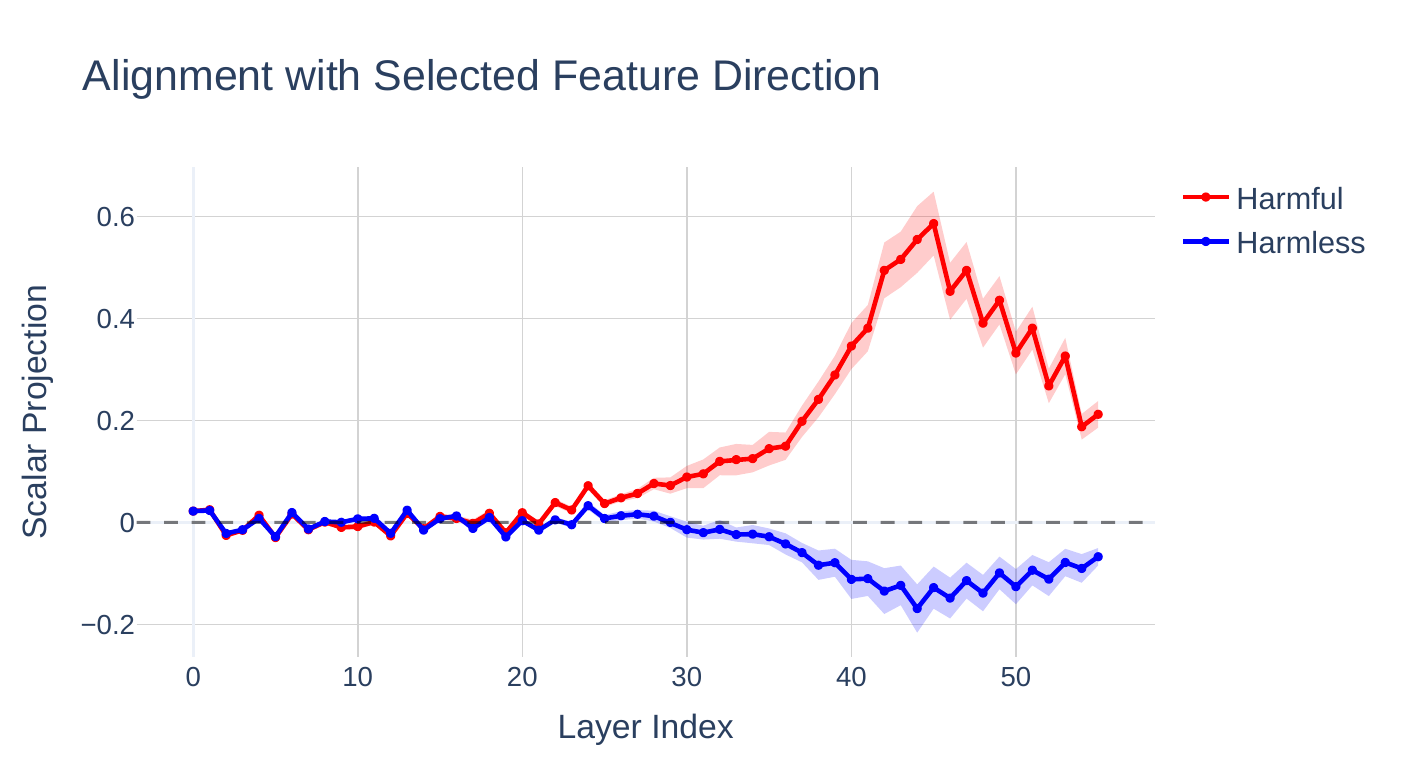}
  }
  \caption{
    \textbf{Layer-wise heterogeneity in Qwen2.5-7B-Instruct.} 
    (a) Activation norms vary substantially across depth, with rapid growth in early layers and amplification near output. 
    (b) Scalar projections class means onto the selected feature direction reveal progressive emergence of opposite-signed discriminability.
  }
  \label{fig:motivation}
\end{figure*}

We analyze activation statistics across model depth using Qwen2.5-7B-Instruct~\citep{qwen2, qwen2.5}. Figure~\ref{fig:motivation} (More in Appendix~\ref{app:heterogeneity_all_models}) reveals two critical phenomena:

\paragraph{Non-uniform Norm Profiles.}
Figure~\ref{fig:activation_norms} shows substantial norm heterogeneity: early layers exhibit rapid growth with high variance, middle layers stabilize, and late layers show dramatic increase near output. Critically, harmful and harmless activations maintain similar norm profiles, motivating examination of \emph{directional properties}.

\paragraph{Progressive Opposite-Signed Discriminability.}
Figure~\ref{fig:projections_local} shows scalar projections of normalized activations onto the chosen direction $\hat{\vd}_{\text{feat}}$, revealing three regimes:
\begin{compactenum}
    \item \textbf{Early layers}: Both classes project near zero with substantial overlap - the feature has not emerged.
    \item \textbf{Middle layers}: Clear separation with opposite-signed projections: harmful samples project positively, harmless negatively. Tight clustering indicates robust discrimination.
    \item \textbf{Late layers}: The separation persists but weakens as the strength decreases.
\end{compactenum}

\paragraph{Key Insight.} Layers where $\boldsymbol{\tilde{\mu}}^{(k)}_{\text{pos}} \cdot \boldsymbol{\tilde{\mu}}^{(k)}_{\text{neg}} < 0$ (opposite-signed mean projections) are optimal steering targets. Uniform steering across all layers disrupts non-discriminative layers, causing coherence collapse.

\subsection{Selective Steering: Norm-Preserving Layer-Wise Control}

\paragraph{Core Innovation.}
We propose \textbf{Selective Steering}, combining: (1) the mathematically sound rotation matrix $\mR^P_\theta$ (Equation~\ref{eq:angular_rotation_matrix}) which inherently preserves norms; (2) selective application only to discriminative layers identified by opposite-signed projections.

\begin{proposition}[Norm Preservation in Selective Steering]
\label{prop:norm_preservation}
The transformation $\vh' = \mR^P_\theta \vh$ preserves norms: $\|\vh'\| = \|\vh\|$ for all $\vh$ and $\theta$, where 

\begin{align}
\label{eq:angular_rotation_matrix}
    \mR^P_\theta = \mI - (\vb_1\vb_1^\top + \vb_2\vb_2^\top) + [\vb_1 \; \vb_2] \, \mR_\theta \,[\vb_1 \; \vb_2]^\top.
\end{align}
\end{proposition}

The proof (Appendix~\ref{app:proof_norm_preservation}) establishes that $\mR^P_\theta$ is an orthogonal transformation by decomposing it into orthogonal projection onto complement space $Q$ and rotation within plane $P$.

\paragraph{Feature Direction Selection.}
Following~\cite{vu2025angular}, we select a global feature direction using difference-in-means with maximum inter-layer consistency. At each layer $k$, compute the local candidate direction:
\begin{align}
\label{eq:local_candidate}
\vd^{(k)} = \boldsymbol{\mu}^{(k)}_{\text{pos}} - \boldsymbol{\mu}^{(k)}_{\text{neg}},
\end{align}
where $\boldsymbol{\mu}^{(k)}_{\text{pos}}$ and $\boldsymbol{\mu}^{(k)}_{\text{neg}}$ are class means from Equation~\ref{eq:diff-in-mean}. The global feature direction is the candidate with highest average cosine similarity to others:
\begin{align}
\label{eq:global_direction}
\hat{\vd}_{\text{feat}} = \text{argmax}_{\vd^{(k)}} \left\{ \frac{1}{L} \sum_{j=1}^{L} \cos(\vd^{(k)}, \vd^{(j)}) \right\},
\end{align}
where $L$ is the number of layers. This selects the direction most consistently represented across depth, capturing the core behavioral axis while filtering layer-specific noise.

\paragraph{Discriminative Layer Selection.}
Given calibration datasets $\mathcal{D}^{(\text{train})}_{\text{pos}}$ and $\mathcal{D}^{(\text{train})}_{\text{neg}}$, we compute mean activations as in Equation~\ref{eq:diff-in-mean}. We define \textbf{discriminative layers}:

\begin{align}
\label{eq:discriminative_layers}
\boldsymbol{\tilde{\mu}}^{(k)}_{\text{pos}} &= \boldsymbol{\mu}^{(k)}_{\text{pos}} \cdot \hat{\vd}_{\text{feat}}, \boldsymbol{\tilde{\mu}}^{(k)}_{\text{neg}} = \boldsymbol{\mu}^{(k)}_{\text{neg}} \cdot \hat{\vd}_{\text{feat}} \notag\\
\mathcal{L}_{\text{disc}} &= \left\{ k \in \{1, \dots, L\} : \boldsymbol{\tilde{\mu}}^{(k)}_{\text{pos}} \cdot \boldsymbol{\tilde{\mu}}^{(k)}_{\text{neg}} < 0 \right\}.
\end{align}

This criterion identifies layers where classes point in opposing directions, ensuring: (1) strong feature representation; (2) predictable steering effect; (3) robust separation across samples.

\paragraph{Steering Transformation.}
For $k \in \mathcal{L}_{\text{disc}}$, we construct a global steering plane $P = \text{span}\{\vb_1, \vb_2\}$ following~\cite{vu2025angular}, where $\vb_1$ is the normalized feature direction and $\vb_2$ is the orthogonalized first principal component of candidate directions. We apply:
\begin{align}
\label{eq:selective_transform}
\vh'^{(k)} = \begin{cases}
\mR^P_{\theta} \vh^{(k)}, & \text{if } k \in \mathcal{L}_{\text{disc}}, \\
\vh^{(k)}, & \text{otherwise},
\end{cases}
\end{align}
where $\mR^P_{\theta} = \mI - (\vb_1\vb_1^\top + \vb_2\vb_2^\top) + [\vb_1 \; \vb_2] \, \mR_{\theta} \,[\vb_1 \; \vb_2]^\top$ and $\mR_{\theta}$ is the 2D rotation matrix. By Proposition~\ref{prop:norm_preservation}, $\|\vh'^{(k)}\| = \|\vh^{(k)}\|$ is guaranteed.

\subsection{Algorithm and Calibration}

Algorithm~\ref{alg:selective_steering} summarizes the inference-time procedure:

\begin{algorithm}[!htbp]
\caption{Selective Steering (Inference)}
\label{alg:selective_steering}
\begin{algorithmic}[1]
\Require Activation $\vh^{(k)}$, basis $\{\vb_1, \vb_2\}$, angle $\theta$, means $\boldsymbol{\mu}^{(k)}_{\text{pos}}, \boldsymbol{\mu}^{(k)}_{\text{neg}}$
\Ensure Steered activation $\vh'^{(k)}$

\If{$\boldsymbol{\tilde{\mu}}^{(k)}_{\text{pos}} \cdot \boldsymbol{\tilde{\mu}}^{(k)}_{\text{neg}} \geq 0$} \Comment{Non-discriminative layer}
    \State \Return $\vh^{(k)}$
\EndIf

\State $\mR_{\theta} \gets \begin{bmatrix} \cos(\theta) & -\sin(\theta) \\ \sin(\theta) & \cos(\theta) \end{bmatrix}$
\State $\mR^P_{\theta} \gets \mI - (\vb_1\vb_1^\top + \vb_2\vb_2^\top) + [\vb_1 \; \vb_2] \, \mR_{\theta} \,[\vb_1 \; \vb_2]^\top$
\State $\vh'^{(k)} \gets \mR^P_{\theta} \vh^{(k)}$ \Comment{Norm preserved by Prop.~\ref{prop:norm_preservation}}
\State \Return $\vh'^{(k)}$
\end{algorithmic}
\end{algorithm}

\paragraph{Calibration.}
One-time setup: (1) extract activations from $\mathcal{D}^{(\text{train})}_{\text{pos}}$ and $\mathcal{D}^{(\text{train})}_{\text{neg}}$; (2) compute $\boldsymbol{\mu}^{(k)}_{\text{pos}}, \boldsymbol{\mu}^{(k)}_{\text{neg}}$ per layer; (3) identify $\mathcal{L}_{\text{disc}}$ via Equation~\ref{eq:discriminative_layers}; (4) construct global plane $P$ via PCA. See Appendix~\ref{app:calibration_details} for full procedure.

\paragraph{Advantages.}
Selective Steering offers: (1) \textbf{guaranteed norm preservation} via Proposition~\ref{prop:norm_preservation}; (2) \textbf{focused intervention} on discriminative layers only; (3) \textbf{reduced computation} from $O(Ld_{\text{model}})$ to $O(|\mathcal{L}_{\text{disc}}|d_{\text{model}})$ where $|\mathcal{L}_{\text{disc}}| \ll L$; (4) \textbf{compatibility} with normalization-heavy architectures.

\section{Experiments}
\label{sec:experiments}

\begin{figure*}[!h]
    \centering
    \includegraphics[width=\linewidth]{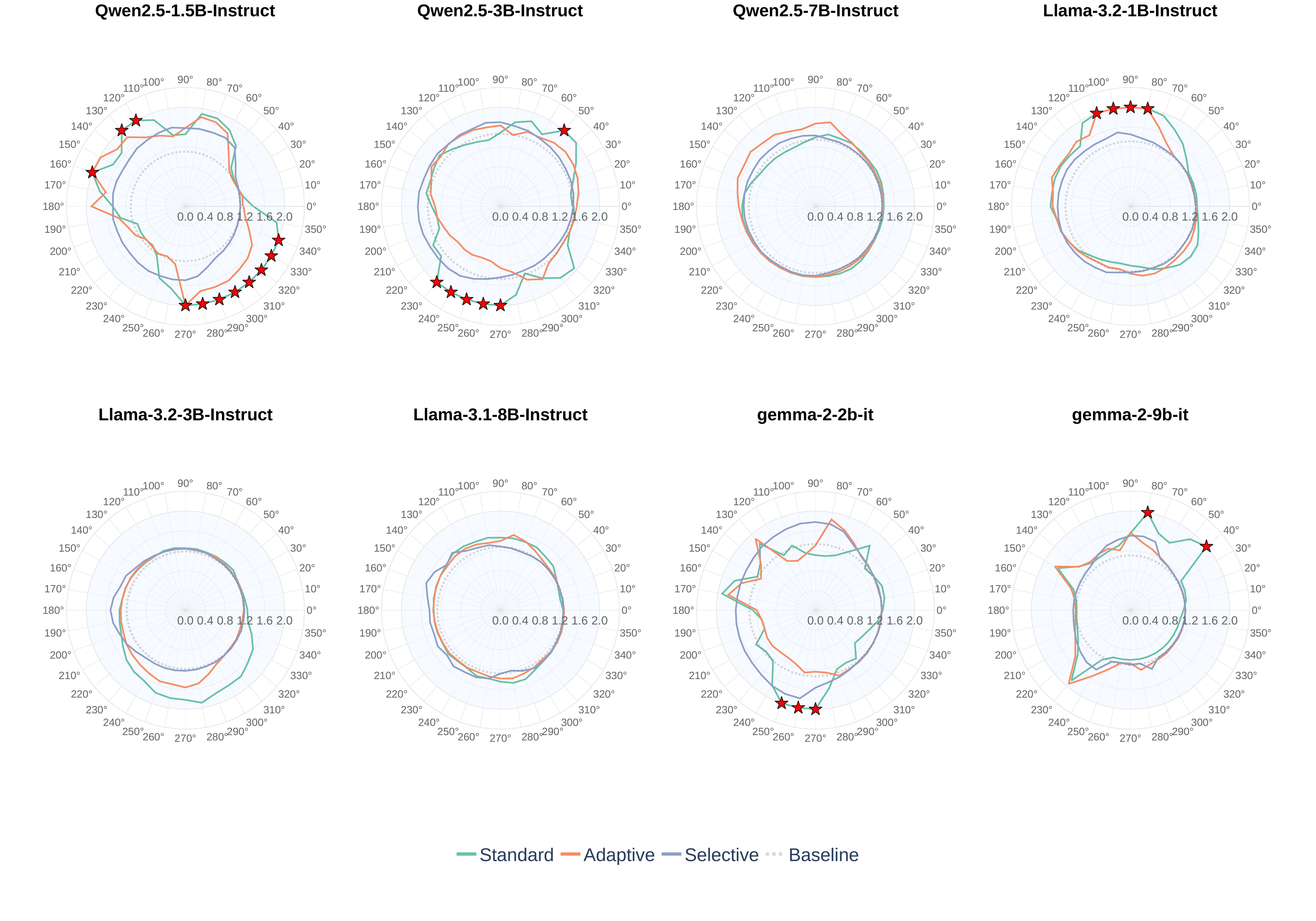}
    \caption{Perplexity measurements across the full steering circle (0°-360°, 10° intervals) for \textbf{SAS, AAS,} and \textbf{Selective Steering (SS)}. Each subplot shows one model's perplexity profile, with the baseline (no steering) shown as a dashed circle. {\color{red}Red stars} indicate angles where perplexity exceeds the threshold of 2.0, signaling generation instability or collapse. \textbf{ActAdd} and \textbf{DirAbl} are excluded as they provide only single-point steering rather than continuous angular control.}
    \label{fig:perplexity}
\end{figure*}

\subsection{Experimental Setup}

\paragraph{Hardware.}
All experiments are conducted on a single NVIDIA A40 GPU with 48GB memory. To ensure reproducibility, we use greedy decoding (temperature = 0.0) across all methods and models. 

\paragraph{Datasets.}
We use two contrastive datasets for calibration: \textbf{AdvBench}~\citep{zou2023universal} (80\%, 416 samples) as $\mathcal{D}_{\text{pos}}^{(\text{train})}$ containing harmful prompts, and 416 samples from \textbf{Alpaca}~\citep{alpaca} as $\mathcal{D}_{\text{neg}}^{(\text{train})}$ containing harmless prompts. The remaining 20\% of AdvBench (104 samples) serves as the evaluation set for measuring coherence and controllability.

To assess robustness, we employ benchmark datasets from \textbf{tinyBenchmarks}~\citep{polo2024tinybenchmarks}, including: tinyAI2\_arc~\citep{clark2018thinksolvedquestionanswering}, tinyGSM8K~\citep{cobbe2021trainingverifierssolvemath}, tinyMMLU~\citep{hendrycks2021measuring}, tinyTruthfulQA~\citep{lin-etal-2022-truthfulqa}, and tinyWinogrande~\citep{sakaguchi2019winogrande}. Each benchmark contains 100 samples.

\paragraph{Baselines.}
We compare against: \textbf{Activation Addition (ActAdd)}~\citep{turner2024steeringlanguagemodelsactivation}, \textbf{Directional Ablation (DirAbl)}~\citep{arditi2024refusal}, \textbf{Standard Angular Steering (SAS)}, and \textbf{Adaptive Angular Steering (AAS)}~\citep{vu2025angular}.

\paragraph{Models.}
We evaluate across three model families with varying sizes: \textbf{Llama}~\citep{grattafiori2024llama3herdmodels} (3.1-8B, 3.2-1B, 3.2-3B), \textbf{Qwen}~\citep{qwen2,qwen2.5} (2.5-1.5B, 2.5-3B, 2.5-7B), and \textbf{Gemma}~\citep{gemmateam2024gemma2improvingopen} (2-2b, 2-9b). All models are instruction-tuned variants trained with alignment data.

\subsection{Evaluation Metrics}

We evaluate Selective Steering across three dimensions: coherence (generation quality), controllability (steering effectiveness), and robustness (capability preservation). Brief metric descriptions are provided below; full mathematical formulations appear in Appendix~\ref{app:metrics}.

\begin{table*}[!h]
\centering
\small
\begin{tabular}{@{}llcccc@{}}
\toprule
\textbf{Model} & \textbf{Method} & \textbf{HarmBench} ↑ & \textbf{PolyGuard} ↑ & \textbf{LLM Judge} ↑ & \textbf{Refusal} ↓ \\
\midrule
\multirow{6}{*}{Llama-3.1-8B}
& ActAdd & \underline{0.7404} & 0.8942 & \underline{0.6827} & \textbf{0.0096} \\
& DirAbl & 0.3269 & 0.3750 & 0.1635 & 0.5288 \\
& SAS & \underline{0.7404} & 0.8942 & \underline{0.6827} & \textbf{0.0096} \\
& AAS & \textbf{0.7788} & \underline{0.9038} & \textbf{0.7019} & \textbf{0.0096} \\
& SS (Ours) & \textbf{0.7788} & \textbf{0.9231} & \textbf{0.7019} & \underline{0.0865} \\
\midrule
\multirow{6}{*}{Llama-3.2-1B}
& ActAdd & 0.7019 & \textbf{0.9904} & 0.7212 & \textbf{0.0000} \\
& DirAbl & 0.5481 & 0.6731 & 0.4423 & \underline{0.2019} \\
& SAS & 0.7019 & \textbf{0.9904} & 0.7212 & \textbf{0.0000} \\
& AAS & \underline{0.7692} & \underline{0.9808} & \underline{0.7308} & \textbf{0.0000} \\
& SS (Ours) & \textbf{0.7981} & \textbf{0.9904} & \textbf{0.7885} & \textbf{0.0000} \\
\midrule
\multirow{6}{*}{Llama-3.2-3B}
& ActAdd & 0.8269 & \underline{0.9519} & \underline{0.8558} & \textbf{0.0000} \\
& DirAbl & 0.5385 & 0.5769 & 0.3654 & \underline{0.2404} \\
& SAS & 0.8269 & \underline{0.9519} & \underline{0.8558} & \textbf{0.0000} \\
& AAS & \underline{0.8462} & \underline{0.9519} & \underline{0.8558} & \textbf{0.0000} \\
& SS (Ours) & \textbf{0.8558} & \textbf{0.9615} & \textbf{0.8654} & \textbf{0.0000} \\
\midrule
\multirow{6}{*}{Qwen2.5-1.5B}
& ActAdd & 0.1346 & \textbf{1.0000} & 0.0385 & \textbf{0.0000} \\
& DirAbl & 0.2500 & 0.3269 & 0.1635 & \underline{0.6250} \\
& SAS & 0.1346 & \textbf{1.0000} & 0.0385 & \textbf{0.0000} \\
& AAS & \underline{0.3942} & \textbf{1.0000} & \underline{0.2981} & \textbf{0.0000} \\
& SS (Ours) & \textbf{0.7404} & \underline{0.9423} & \textbf{0.6635} & \textbf{0.0000} \\
\midrule
\multirow{6}{*}{Qwen2.5-3B}
& ActAdd & 0.5096 & \textbf{1.0000} & 0.2885 & \textbf{0.0000} \\
& DirAbl & 0.5288 & 0.6442 & 0.4327 & \underline{0.0192} \\
& SAS & 0.5096 & \textbf{1.0000} & 0.2885 & \textbf{0.0000} \\
& AAS & \underline{0.7019} & \textbf{1.0000} & \underline{0.5673} & \textbf{0.0000} \\
& SS (Ours) & \textbf{0.8462} & \underline{0.9615} & \textbf{0.8365} & \textbf{0.0000} \\
\midrule
\multirow{6}{*}{Qwen2.5-7B}
& ActAdd & \underline{0.8654} & \textbf{0.9904} & \textbf{0.9038} & \textbf{0.0000} \\
& DirAbl & 0.5577 & 0.6538 & 0.4712 & \underline{0.0577} \\
& SAS & \underline{0.8654} & \textbf{0.9904} & \textbf{0.9038} & \textbf{0.0000} \\
& AAS & \textbf{0.8750} & \underline{0.9712} & \underline{0.8750} & \textbf{0.0000} \\
& SS (Ours) & \textbf{0.8750} & 0.9423 & 0.8173 & \textbf{0.0000} \\
\midrule
\multirow{6}{*}{gemma-2-2b}
& ActAdd & 0.0000 & \textbf{1.0000} & 0.0000 & \textbf{0.0000} \\
& DirAbl & 0.2500 & 0.3462 & 0.2404 & \underline{0.0192} \\
& SAS & 0.0000 & \textbf{1.0000} & 0.0000 & \textbf{0.0000} \\
& AAS & \underline{0.7404} & \textbf{1.0000} & \underline{0.7212} & \textbf{0.0000} \\
& SS (Ours) & \textbf{0.8269} & \underline{0.9712} & \textbf{0.8269} & \textbf{0.0000} \\
\midrule
\multirow{6}{*}{gemma-2-9b}
& ActAdd & 0.0000 & \textbf{1.0000} & 0.0000 & \textbf{0.0000} \\
& DirAbl & 0.1154 & \underline{0.1538} & 0.0962 & \underline{0.0769} \\
& SAS & 0.0000 & \textbf{1.0000} & 0.0000 & \textbf{0.0000} \\
& AAS & \underline{0.6731} & \textbf{1.0000} & \underline{0.5096} & \textbf{0.0000} \\
& SS (Ours) & \textbf{0.6827} & \textbf{1.0000} & \textbf{0.6827} & \textbf{0.0000} \\
\bottomrule
\end{tabular}%
\caption{Controllability evaluation at best steering per method. Best scores (excluding No Steering) in \textbf{bold}, second-best \underline{underlined}.}
\label{tab:controllability}
\end{table*}

\paragraph{Coherence Metrics.}
We employ four complementary metrics:
\begin{compactenum}
    \item \textbf{Perplexity (PPL↓):} Measures model uncertainty. Lower indicates more confident generation.
    \item \textbf{N-gram Repetition (N-gram Rep.↓):} Detects pathological repetition using 4-gram diversity. Lower indicates less repetition.
    \item \textbf{Language Consistency (Lang. Cons.↑):} Detects foreign character contamination via Unicode script analysis. Higher indicates fewer unwanted script intrusions.
    \item \textbf{Compression Ratio (Comp. Ratio↑):} Pattern-agnostic collapse detection using gzip. Higher indicates more diverse, natural text.
\end{compactenum}

\paragraph{Controllability Metrics.}
We measure steering effectiveness using:
\begin{compactenum}
    \item \textbf{Attack Success Rate (ASR↑):} Proportion of harmful prompts eliciting harmful responses, evaluated using three classifiers: HarmBench~\citep{mazeika2024harmbench}, PolyGuard~\citep{kumar2025polyguard}, and LLM-as-judge with Qwen2.5-14B-Instruct~\citep{qwen2.5}. Higher indicates more successful steering.
    \item \textbf{Refusal Score (RS↓)~\citep{arditi2024refusal}:} Substring-based detection of refusal patterns (e.g., "I'm sorry", "I cannot"). Lower indicates less refusal behavior.
\end{compactenum}

\paragraph{Robustness Metrics.}
We measure general capability preservation using:
\begin{compactenum}
    \item \textbf{Accuracy (Acc↑):} Zero-shot accuracy on tinyBenchmarks suite~\citep{polo2024tinybenchmarks}. Higher indicates better capability retention.
\end{compactenum}

Arrows (↑/↓) indicate whether higher or lower values are better.

\begin{table*}[!ht]
\centering
\small
\begin{tabular}{@{}llcccccc@{}}
\toprule
\textbf{Model} & \textbf{Method} & \textbf{ASR} ↑ & \textbf{AI2\_arc} & \textbf{GSM8k} & \textbf{MMLU} & \textbf{TruthfulQA} & \textbf{Winogrande} \\
\midrule
\rowcolor{gray!30}
\multirow{6}{*}{Llama-3.1-8B}
& No Steering & 0.0577 & 0.8100 & 0.8500 & 0.6600 & 0.5600 & 0.5100 \\
& ActAdd & 0.7404 & 0.6100 & 0.6400 & 0.5100 & 0.3900 & 0.3500 \\
& DirAbl & 0.3269 & \textbf{0.8000} & \underline{0.8600} & \textbf{0.6700} & \underline{0.5600} & 0.4900 \\
& SAS & 0.7404 & 0.6100 & 0.6400 & 0.5100 & 0.3900 & 0.3500 \\
& AAS & \underline{0.7788} & \underline{0.7700} & \textbf{0.8800} & \textbf{0.6700} & \textbf{0.5700} & 0.4700 \\
& SS (Ours) & \textbf{0.7788} & \textbf{0.8000} & \textbf{0.8800} & \underline{0.6600} & 0.5500 & \textbf{0.5100} \\
\midrule
\rowcolor{gray!30}
\multirow{6}{*}{Llama-3.2-1B}
& No Steering & 0.0673 & 0.4700 & 0.4300 & 0.4600 & 0.2100 & 0.3100 \\
& ActAdd & 0.7019 & 0.1700 & 0.1200 & 0.0700 & 0.0300 & 0.0200 \\
& DirAbl & 0.5481 & 0.4100 & \underline{0.4000} & 0.3800 & 0.3100 & \underline{0.3500} \\
& SAS & 0.7019 & 0.1700 & 0.1200 & 0.0700 & 0.0300 & 0.0200 \\
& AAS & \underline{0.7692} & \underline{0.4500} & 0.3500 & 0.4200 & \underline{0.2000} & \textbf{0.3600} \\
& SS (Ours) & \textbf{0.7981} & \textbf{0.4600} & \textbf{0.4600} & \textbf{0.4200} & \textbf{0.2200} & 0.3100 \\
\midrule
\rowcolor{gray!30}
\multirow{6}{*}{Llama-3.2-3B}
& No Steering & 0.0192 & 0.7100 & 0.8000 & 0.6100 & 0.5700 & 0.3600 \\
& ActAdd & \underline{0.8269} & 0.4100 & 0.6800 & 0.3300 & 0.3900 & 0.3600 \\
& DirAbl & 0.5385 & 0.6700 & 0.7500 & \textbf{0.6100} & \textbf{0.5900} & 0.3400 \\
& SAS & \underline{0.8269} & 0.2400 & 0.4600 & 0.1500 & 0.2000 & 0.2900 \\
& AAS & 0.8462 & \underline{0.7000} & \textbf{0.8100} & \underline{0.5900} & 0.5600 & \textbf{0.4200} \\
& SS (Ours) & \textbf{0.8558} & \textbf{0.7200} & \underline{0.7800} & \textbf{0.6100} & \underline{0.5700} & \underline{0.3700} \\
\midrule
\rowcolor{gray!30}
\multirow{6}{*}{Qwen2.5-1.5B}
& No Steering & 0.0000 & 0.6900 & 0.7800 & 0.5300 & 0.4900 & 0.4700 \\
& ActAdd & 0.1346 & 0.0800 & 0.0000 & 0.0600 & 0.1800 & 0.1000 \\
& DirAbl & 0.2500 & 0.6600 & \textbf{0.7600} & 0.4800 & 0.4300 & 0.4300 \\
& SAS & 0.1346 & 0.0800 & 0.0000 & 0.0800 & 0.3700 & 0.1700 \\
& AAS & \underline{0.3942} & \underline{0.7000} & \underline{0.7200} & \underline{0.5000} & \textbf{0.5100} & \underline{0.4500} \\
& SS (Ours) & \textbf{0.7404} & \textbf{0.6900} & \underline{0.7200} & \textbf{0.5200} & \underline{0.4800} & \textbf{0.4700} \\
\midrule
\rowcolor{gray!30}
\multirow{6}{*}{Qwen2.5-3B}
& No Steering & 0.0000 & 0.8000 & 0.8800 & 0.6100 & 0.6000 & 0.5300 \\
& ActAdd & 0.5096 & 0.0100 & 0.0000 & 0.0000 & 0.0000 & 0.0000 \\
& DirAbl & 0.5288 & \textbf{0.8000} & 0.8200 & \textbf{0.6200} & \underline{0.5700} & \underline{0.5000} \\
& SAS & 0.5096 & 0.0100 & 0.0000 & 0.0000 & 0.0000 & 0.0000 \\
& AAS & \underline{0.7019} & 0.7800 & \underline{0.8500} & 0.5200 & 0.3400 & \underline{0.5000} \\
& SS (Ours) & \textbf{0.8462} & \underline{0.7900} & \textbf{0.8800} & \underline{0.6100} & \textbf{0.6100} & \textbf{0.5300} \\
\midrule
\rowcolor{gray!30}
\multirow{6}{*}{Qwen2.5-7B}
& No Steering & 0.0000 & 0.8700 & 0.9300 & 0.6400 & 0.6300 & 0.5900 \\
& ActAdd & \underline{0.8654} & 0.7900 & 0.8100 & \underline{0.6800} & 0.3600 & 0.4900 \\
& DirAbl & 0.5577 & 0.8600 & \underline{0.9200} & \textbf{0.6400} & \underline{0.5700} & \textbf{0.6100} \\
& SAS & \underline{0.8654} & 0.7900 & 0.8100 & \underline{0.6800} & 0.3600 & 0.4900 \\
& AAS & \textbf{0.8750} & \textbf{0.9000} & 0.9100 & \textbf{0.6900} & 0.4700 & 0.4500 \\
& SS (Ours) & \underline{0.8750} & \textbf{0.8700} & \textbf{0.9400} & 0.6500 & \textbf{0.6300} & \underline{0.5900} \\
\midrule
\rowcolor{gray!30}
\multirow{6}{*}{gemma-2-2b}
& No Steering & 0.0000 & 0.7100 & 0.7000 & 0.5400 & 0.5500 & 0.3800 \\
& ActAdd & 0.0000 & 0.0000 & 0.0000 & 0.0000 & 0.0100 & 0.0000 \\
& DirAbl & 0.2500 & \textbf{0.7300} & \underline{0.6500} & \textbf{0.5600} & \textbf{0.5800} & \textbf{0.4300} \\
& SAS & 0.0000 & 0.0000 & 0.0000 & 0.0000 & 0.0100 & 0.0000 \\
& AAS & \underline{0.7404} & 0.3800 & 0.0800 & 0.1300 & 0.1400 & 0.2700 \\
& SS (Ours) & \textbf{0.8269} & \underline{0.7100} & \textbf{0.6900} & \underline{0.5400} & \underline{0.5600} & \underline{0.4000} \\
\midrule
\rowcolor{gray!30}
\multirow{6}{*}{gemma-2-9b}
& No Steering & 0.0000 & 0.9000 & 0.9300 & 0.7100 & 0.7400 & 0.5900 \\
& ActAdd & 0.0000 & \underline{0.0000} & 0.0000 & 0.0000 & 0.0000 & 0.0000 \\
& DirAbl & 0.1154 & \textbf{0.9000} & \textbf{0.9400} & 0.7000 & \underline{0.7400} & \textbf{0.5900} \\
& SAS & 0.0000 & \underline{0.0000} & 0.0000 & 0.0000 & 0.0000 & 0.0000 \\
& AAS & \underline{0.6731} & \textbf{0.9000} & \underline{0.9300} & \textbf{0.7200} & \textbf{0.7500} & \underline{0.5700} \\
& SS (Ours) & \textbf{0.6827} & \textbf{0.9000} & \underline{0.9300} & \underline{0.7100} & \underline{0.7400} & \textbf{0.5900} \\
\bottomrule
\end{tabular}%
\caption{Robustness evaluation on tinyBenchmarks at best HarmBench ASR angle per method. Best scores (excluding No Steering) in \textbf{bold}, second-best \underline{underlined}.}
\label{tab:robustness}
\end{table*}

\subsection{Results}
\paragraph{Coherence Analysis.}
Figure~\ref{fig:perplexity} presents perplexity measurements across the steering circle for SAS, AAS, and SS. Red stars indicate angles where perplexity exceeds the threshold (default: 2.0), signaling potential generation collapse. \textbf{SS demonstrates remarkably stable perplexity across all angles and models}, with zero threshold violations across 8 models. In contrast, SAS and AAS exhibit frequent spikes, particularly in smaller models (Llama-3.2-1B, Qwen2.5-1.5B, gemma-2-2b) and at critical angles (80°-160°, 220°-350°). Table~\ref{tab:coherence} quantifies coherence quality through three complementary metrics. \textbf{SS achieves the best or second-best compression ratio in 8/8 models}, indicating superior resistance to generation collapse (More in Appendix~\ref{app:additional-results}). 


\paragraph{Controllability Analysis.}
Table~\ref{tab:controllability} evaluates steering effectiveness using multiple ASR metrics, the most challenging benchmark. \textbf{SS achieves the highest or second-highest ASR in 8/8 models on HarmBench}. Critically, \textbf{SS demonstrates superior controllability on smaller and harder-to-steer models}: on Qwen2.5-1.5B, SS achieves 74.04\% HarmBench ASR versus 39.42\% for AAS and 13.46\% for SAS - a \textbf{5.5× improvement over SAS}. On gemma-2-2b, where SAS completely fails (0\% ASR) and AAS achieves only 74.04\%, \textbf{SS reaches 82.69\% ASR}.

The refusal score metric reveals SS maintains lower refusal rates comparable to other methods, with 0\% refusal in 7/8 models. Notably, SS balances high ASR with consistent performance across all three evaluators (HarmBench, PolyGuard, LLM-judge), avoiding the specialized overfitting seen in some baselines.

\paragraph{Robustness Analysis.}
Table~\ref{tab:robustness} evaluates zero-shot performance on general capabilities benchmarks at each method's best ASR steering angle. \textbf{SS preserves baseline performance significantly better than competing methods}, achieving the best or second-best average accuracy across benchmarks and models.

The robustness advantage is most pronounced on models where steering poses challenges. On Qwen2.5-3B, SAS again causes complete collapse (0.88→0.00 on tinyGSM8K), whereas \textbf{SS preserves 100\% of baseline (0.88→0.88)}. On gemma-2-2b/9b, where ActAdd and SAS produce degenerate outputs (0\% across all benchmarks), \textbf{SS maintains approximately 100\% of baseline performance}.

Notably, SS achieves this robustness \emph{without sacrificing controllability}: on Qwen2.5-3B, SS simultaneously delivers 84.62\% HarmBench ASR (highest among all methods) and maintains benchmark accuracy. This demonstrates that \textbf{selective layer intervention successfully decouples steering effectiveness from general capability preservation}.

\paragraph{Summary.}
Across three comprehensive evaluation dimensions, \textbf{Selective Steering (SS) consistently outperforms existing methods by simultaneously achieving: (1) superior generation coherence with zero perplexity threshold violations, (2) state-of-the-art controllability especially on challenging small models (up to 5.5× improvement), and (3) near-perfect preservation of general capabilities (approximately 100\% baseline retention)}. The combination of norm-preserving rotation and discriminative layer selection enables robust, effective steering without the catastrophic degradation observed in SAS/AAS or the collapse-prone behavior of ActAdd on certain model families.

\section{Conclusion}
\label{sec:conclusion}

We presented \textbf{Selective Steering}, a principled activation steering method that achieves robust, controllable behavior modification in large language models through two complementary innovations: norm-preserving rotation and discriminative layer selection.

Our theoretical analysis (Propositions~\ref{prop:angular_violation} and~\ref{prop:norm_preservation}) establishes that prior rotation-based steering suffers from fundamental norm violations, causing distribution shift that prevents effective control, especially in smaller models. By adopting the mathematically sound rotation matrix formulation, Selective Steering guarantees $\|\vh'\| = \|\vh\|$, eliminating coherence collapse while enabling precise angular control.

Empirically, we demonstrated that feature discriminability - measured by opposite-signed mean projections $\boldsymbol{\mu}^{(k)}_{\text{pos}} \cdot \boldsymbol{\mu}^{(k)}_{\text{neg}} < 0$ - emerges progressively across model depth, concentrating in specific middle layers. By restricting intervention to these discriminative layers ($\mathcal{L}_{\text{disc}}$), Selective Steering focuses steering effect where features are most strongly represented, avoiding interference in non-discriminative regions.

Comprehensive experiments across nine models spanning 1.5B to 9B parameters validate our approach. Selective Steering achieves 5.5$\times$ higher attack success rates than Angular Steering and Adaptive Angular Steering, with zero perplexity violations and approximately 100\% accuracy retention on 5 standard benchmarks. Ablation studies confirm that both norm preservation and discriminative layer selection are essential: removing either component causes dramatic performance degradation.



\section{Limitations}
\label{sec:limitations}

While Selective Steering demonstrates strong empirical performance, our approach inherits limitations from its methodological foundations:

\textbf{Feature Direction Extraction.} Following prior work \citep{arditi2024refusal, turner2024steeringlanguagemodelsactivation, zou2025representation}, we use difference-in-means to extract feature directions. While simple and effective, this approach is not guaranteed to identify the optimal discriminative direction. More sophisticated methods such as Fisher discriminant analysis, or sparse dictionary learning \citep{templeton2024scaling} may yield superior directions, though at increased computational cost. Our discriminative layer selection criterion ($\mu^{(k)}_{\text{pos}} \cdot \mu^{(k)}_{\text{neg}} < 0$) naturally extends to any feature extraction method.

\textbf{Steering Plane Construction.} Our 2D plane construction combines the selected feature direction with the first principal component from PCA over candidate directions - a heuristic also used in Angular Steering \citep{vu2025angular}. While this captures the primary variance in layer-wise feature evolution, it lacks theoretical guarantees for optimality. Alternative constructions using the second-best discriminative direction, orthogonal basis optimization \citep{pham-nguyen-2024-householder}, or Grassmannian manifold methods may improve steering effectiveness. Despite this heuristic nature, our empirical results demonstrate that the current construction is sufficient for robust control across diverse model families and sizes.

These limitations represent opportunities for future refinement rather than fundamental flaws, as our core contributions - discriminative layer selection and norm preservation - remain valid regardless of the specific feature extraction or plane construction method employed.

\section*{Ethics Statement} \label{sec:ethics}
\addcontentsline{toc}{section}{Ethics Statement}
The development of Selective Steering is motivated by the need to understand and control large language model (LLM) behaviors, particularly in safety-critical contexts such as content moderation and harmful request refusal. We recognize the dual-use nature of activation steering techniques: while they enable beneficial applications like improving model alignment and robustness, they could potentially be misused to bypass safety mechanisms or manipulate model outputs in harmful ways.

To address these concerns, our research is conducted with a commitment to responsible disclosure and ethical AI development. The steering methods and experimental protocols presented in this work are designed explicitly for diagnostic and improvement purposes - to assess model vulnerabilities, understand internal representations of safety-relevant features, and develop more robust control mechanisms. All experiments involving harmful prompts use established benchmarks that are already publicly available for red-teaming research, and our evaluations measure refusal behavior rather than generating actual harmful content.

We emphasize that Selective Steering, like other activation steering methods, requires direct access to model internals and cannot be applied to API-only deployments, limiting potential misuse vectors. Furthermore, our ablation studies and detailed analysis reveal the conditions under which steering succeeds or fails, providing model developers with insights to develop more resilient architectures and safety mechanisms that are resistant to activation-based manipulation.

The open release of our methodology and code is intended to foster collaborative advances in LLM safety and interpretability within the research community. We encourage researchers and practitioners to use these techniques responsibly: (1) for improving model alignment and safety rather than circumventing protections, (2) in collaboration with model developers to address identified vulnerabilities, (3) with appropriate institutional oversight and ethical review, and (4) in adherence to legal and ethical standards governing AI safety research.

By advancing our understanding of how behavioral features are represented and can be controlled in LLMs, we aim to contribute to the development of more transparent, interpretable, and trustworthy AI systems. We believe that openly studying these mechanisms - including their limitations and failure modes - is essential for building robust safety measures that can withstand adversarial pressures in real-world deployments.


\bibliography{main}

\newpage
\appendix

\section{Related Work}
\label{sec:related}

\subsection{Alignment and Safety in LLMs}

Traditional approaches to LLM safety rely on alignment training through RLHF~\citep{ouyang2022training, bai2022training} and constitutional AI~\citep{bai2022constitutional}, which optimize models to refuse harmful requests while maintaining helpfulness. However, these methods require expensive retraining~\citep{casper2023open}, suffer from reward hacking~\citep{gao2022scaling}, and remain vulnerable to adversarial attacks~\citep{zou2023universal, wei2023jailbroken}. Recent work reveals that alignment creates superficial refusal behaviors rather than removing harmful knowledge~\citep{arditi2024refusal}, motivating inference-time intervention approaches that directly modify model representations.

\subsection{Activation Steering Methods}

\paragraph{Vector Addition Approaches.}
Early steering methods manipulate activations through vector arithmetic. \textbf{Activation Addition}~\citep{turner2024steeringlanguagemodelsactivation} adds scaled feature directions extracted via contrastive mean differences: $h' = h + \alpha d_{\text{feat}}$, where $\alpha$ controls steering intensity. \textbf{Contrastive Activation Addition (CAA)}~\citep{rimsky-etal-2024-steering} extends this with multiple contrastive pairs for robust direction extraction. However, these methods are highly sensitive to coefficient tuning - inappropriate $\alpha$ values cause incoherent generation due to norm distortion~\citep{templeton2024scaling}. Moreover, $\alpha$ must be layer-specific to account for exponentially growing activation norms across depth, making manual tuning impractical.

\paragraph{Subspace Projection Methods.}
\textbf{Directional Ablation (DirAbl)}~\citep{arditi2024refusal} removes features by orthogonal projection: $h' = h - (d_{\text{feat}} \cdot h)d_{\text{feat}}$, eliminating refusal directions entirely. \textbf{Representation Engineering}~\citep{zou2023transparency} generalizes this framework for reading and controlling model representations. While these methods avoid hyperparameter sensitivity, they offer only binary control - features are either fully removed or left intact, precluding fine-grained modulation. Recent work on fairness~\citep{li2025fairsteer} applies similar projection-based interventions but faces the same limitations.

\paragraph{Geometric Rotation Methods.}
\textbf{Standard Angular Steering (SAS)}~\citep{vu2025angular} reformulates steering as norm-preserving rotation within a 2D plane spanned by the feature direction and its principal component. By rotating activations to target angles $\theta$, it provides continuous control and generalizes both addition ($\theta < 180°$) and ablation ($\theta = 90°$). \textbf{Adaptive Angular Steering (AAS)}~\citep{vu2025angular} adds conditional masking, applying rotation only to activations aligned with the feature direction: mask $= \max(0, \text{sign}(h \cdot d_{\text{feat}}))$. However, both methods apply steering uniformly across all layers, causing generation collapse on smaller models and poor controllability on strongly aligned models. Our analysis reveals this stems from ignoring layer-wise discriminability - early layers lack meaningful feature separation while steering them disrupts unrelated representations.

\subsection{Layer-Specific Interventions}

Recent work recognizes layers play heterogeneous roles. \textbf{Circuit analysis}~\citep{wang2023interpretability, marks2024sparse} identifies specific attention heads and MLP neurons responsible for behaviors, enabling surgical interventions. \textbf{Mechanistic interpretability}~\citep{elhage2021mathematical, nanda2023progress} studies information flow through layer-wise transformations, revealing that features emerge progressively across depth. However, these approaches focus on understanding rather than control. Concurrent work on \textbf{layer-wise steering}~\citep{harrasse2025disentangling} observes varying steering effectiveness across layers but lacks principled selection criteria. Our discriminative criterion $\mu_{\text{pos}}^{(k)} \cdot \mu_{\text{neg}}^{(k)} < 0$ provides a theoretically grounded, automatically computable condition for identifying steerable layers.

\subsection{Comparison with Prior Methods}

Table~\ref{tab:method_comparison} contrasts Selective Steering with prior angular methods. Unlike Angular and Adaptive Angular Steering, which violate norm preservation during plane projection (Proposition~\ref{prop:angular_violation}), SS guarantees norm preservation through discriminative layer selection (Proposition~\ref{prop:norm_preservation}). Our opposition-based criterion identifies layers where classes exhibit opposite-signed projections, concentrating steering effort where features naturally separate. This reduces computational overhead from $O(Ld_{\text{model}})$ to $O(|\mathcal{L}_{\text{disc}}|d_{\text{model}})$ where $|\mathcal{L}_{\text{disc}}| \ll L$, as only discriminative layers require rotation matrices.

\begin{table*}[!htbp]
\centering
\caption{Comparison of steering methods on key properties. \ding{51} indicates satisfaction, \ding{55} indicates violation.}
\label{tab:method_comparison}
\resizebox{\textwidth}{!}{%
\begin{tabular}{@{}lccccc@{}}
\toprule
\textbf{Property} & \textbf{ActAdd} & \textbf{DirAbl} & \textbf{SAS} & \textbf{AAS} & \textbf{SS (Ours)} \\
\midrule
Norm preservation & \ding{55} & \ding{55} & \ding{55} & \ding{55} & \ding{51} \\
Layer selectivity & \ding{55} & \ding{55} & \ding{55} & \ding{55} & \ding{51} \\
Continuous control & \ding{55} & \ding{55} & \ding{51} & \ding{51} & \ding{51} \\
Fine-grained modulation & \ding{51} & \ding{55} & \ding{51} & \ding{51} & \ding{51} \\
Discriminability criterion & None & None & None & Alignment & Opposition \\
Hyperparameter sensitivity & High & Low & Low & Low & Low \\
Computational cost & $O(Ld_{\text{model}})$ & $O(Ld_{\text{model}})$ & $O(Ld_{\text{model}})$ & $O(Ld_{\text{model}})$ & $O(|\mathcal{L}_{\text{disc}}|d_{\text{model}})$ \\
\bottomrule
\end{tabular}%
}
\end{table*}

Our method is the first to combine continuous angular control with principled layer selection, achieving robust steering without coherence degradation.
\section{Detailed Methodology}
\label{app:detailed_methodology}

\subsection{Proof: Norm Violation in Angular Steering}
\label{app:proof_angular_violation}

\begin{proof}[Proof of Proposition~\ref{prop:angular_violation}]
We demonstrate a counterexample at the identity case $\theta = 0°$, where intuitively no transformation should occur. For $\theta = 0$, the rotation matrix is:
\begin{align}
\mR_{0} = \begin{bmatrix} 1 & 0 \\ 0 & 1 \end{bmatrix}, \quad \text{thus} \quad \mR_{0} \begin{bmatrix} 1 \\ 0 \end{bmatrix} = \begin{bmatrix} 1 \\ 0 \end{bmatrix}.
\end{align}

Substituting $\theta = 0$ into Equation~\ref{eq:angular_steering_transform}:
\begin{align}
\vh_{\text{steered}, 0}^{\text{AS}} &= \vh - \text{proj}_P(\vh) + \|\text{proj}_P(\vh)\| \cdot [\vb_1 \; \vb_2] \begin{bmatrix} 1 \\ 0 \end{bmatrix} \nonumber \\
&= \vh - \text{proj}_P(\vh) + \|\text{proj}_P(\vh)\| \cdot \vb_1.
\end{align}

For $\vh_{\text{steered}, 0}^{\text{AS}} = \vh$ (identity), we require:
\begin{align}
\label{eq:identity_condition}
- \text{proj}_P(\vh) + \|\text{proj}_P(\vh)\| \cdot \vb_1 = \mathbf{0}.
\end{align}

Let $\text{proj}_P(\vh) = c_1 \vb_1 + c_2 \vb_2$ where $c_1 = \vb_1^\top \vh$ and $c_2 = \vb_2^\top \vh$. Then:
\begin{align}
\|\text{proj}_P(\vh)\| = \sqrt{c_1^2 + c_2^2}.
\end{align}

Substituting into Equation~\ref{eq:identity_condition}:
\begin{align}
-(c_1 \vb_1 + c_2 \vb_2) + \sqrt{c_1^2 + c_2^2} \cdot \vb_1 = \mathbf{0}.
\end{align}

Rearranging:
\begin{align}
\left(\sqrt{c_1^2 + c_2^2} - c_1\right) \vb_1 - c_2 \vb_2 = \mathbf{0}.
\end{align}

Since $\{\vb_1, \vb_2\}$ are orthonormal, both coefficients must vanish:
\begin{align}
\sqrt{c_1^2 + c_2^2} - c_1 = 0 \quad \text{and} \quad c_2 = 0.
\end{align}

Combined with $c_2 = 0$, the first condition simplifies to $|c_1| = c_1$, requiring $c_1 \geq 0$.

\textbf{Thus, $\vh_{\text{steered}, 0}^{\text{AS}} = \vh$ holds only when $\vh$'s projection lies exactly along $\vb_1$ with non-negative coefficient} ($c_2 = 0$ and $c_1 \geq 0$). For general $\vh$ where $c_2 \neq 0$ or $c_1 < 0$:
\begin{align}
\vh_{\text{steered}, 0}^{\text{AS}} \neq \vh \quad \Rightarrow \quad \|\vh_{\text{steered}, 0}^{\text{AS}}\| \neq \|\vh\|.
\end{align}

This demonstrates fundamental norm violation even at the identity transformation.
\end{proof}

\subsection{Proof: Norm Preservation in Selective Steering}
\label{app:proof_norm_preservation}

\begin{proof}[Proof of Proposition~\ref{prop:norm_preservation}]
The rotation matrix decomposes as:
\begin{align}
\mR^P_\theta = \underbrace{[\mI - (\vb_1\vb_1^\top + \vb_2\vb_2^\top)]}_{\text{projection onto } Q} + \underbrace{[\vb_1 \; \vb_2] \, \mR_{\theta} \,[\vb_1 \; \vb_2]^\top}_{\text{rotation in plane } P},
\end{align}
where $Q$ is the orthogonal complement of $P = \text{span}\{\vb_1, \vb_2\}$.

\noindent Decompose $\vh = \vh_P + \vh_Q$ where:
\begin{align}
\vh_P &= (\vb_1\vb_1^\top + \vb_2\vb_2^\top)\vh = c_1\vb_1 + c_2\vb_2, \\
\vh_Q &= [\mI - (\vb_1\vb_1^\top + \vb_2\vb_2^\top)]\vh.
\end{align}

\noindent Applying $\mR^P_\theta$:
\begin{align}
\mR^P_\theta \vh &= [\mI - (\vb_1\vb_1^\top + \vb_2\vb_2^\top)](\vh_P + \vh_Q) \\
&+ [\vb_1 \; \vb_2] \, \mR_{\theta} \,[\vb_1 \; \vb_2]^\top(\vh_P + \vh_Q) \nonumber \\
&= \vh_Q + [\vb_1 \; \vb_2] \, \mR_{\theta} \, [c_1 \; c_2]^\top,
\end{align}
since projection annihilates $\vh_P$, preserves $\vh_Q$, and $[\vb_1 \; \vb_2]^\top \vh_Q = \mathbf{0}$.

\noindent The 2D rotation matrix $\mR_\theta$ is orthogonal: $\mR_\theta^\top \mR_\theta = \mI_2$. Therefore:
\begin{align}
\|\mR^P_\theta \vh\|^2 &= \|\vh_Q\|^2 + \|[\vb_1 \; \vb_2] \, \mR_{\theta} \, [c_1 \; c_2]^\top\|^2 \nonumber \\
&= \|\vh_Q\|^2 + \|\mR_{\theta} \, [c_1 \; c_2]^\top\|^2 \\
& \quad \text{($\{\vb_1, \vb_2\}$ orthonormal)} \nonumber \\
&= \|\vh_Q\|^2 + \|[c_1 \; c_2]^\top\|^2 \\
& \quad \text{($\mR_\theta$ preserves norms)} \nonumber \\
&= \|\vh_Q\|^2 + c_1^2 + c_2^2 \\
&= \|\vh_Q\|^2 + \|\vh_P\|^2 \\
&= \|\vh\|^2,
\end{align}
where the last equality follows from orthogonality of $P$ and $Q$. Thus $\|\mR^P_\theta \vh\| = \|\vh\|$.
\end{proof}

\subsection{Calibration Procedure}
\label{app:calibration_details}

\paragraph{Step 1: Activation Extraction.}
Pass all prompts in $\mathcal{D}^{(\text{train})}_{\text{pos}}$ and $\mathcal{D}^{(\text{train})}_{\text{neg}}$ through the model. At each layer $k \in \{1, \dots, L\}$ (specifically, after normalization before attention and MLP blocks), record the final token's activation vector $\vh^{(k)}_p$ for each prompt $p$.

\paragraph{Step 2: Mean Vector Computation.}
For each layer $k$:
\begin{align}
\boldsymbol{\mu}^{(k)}_{\text{pos}} = \frac{1}{|\mathcal{D}^{(\text{train})}_{\text{pos}}|} 
    \sum_{p \in \mathcal{D}^{(\text{train})}_{\text{pos}}} \vh^{(k)}_p, \\
\boldsymbol{\mu}^{(k)}_{\text{neg}} = \frac{1}{|\mathcal{D}^{(\text{train})}_{\text{neg}}|} 
    \sum_{p \in \mathcal{D}^{(\text{train})}_{\text{neg}}} \vh^{(k)}_p.
\end{align}

\paragraph{Step 3: Global Feature Direction Selection.}
Compute candidate directions at each layer using difference-in-means:
\begin{align}
\vd^{(k)} = \boldsymbol{\mu}^{(k)}_{\text{pos}} - \boldsymbol{\mu}^{(k)}_{\text{neg}}, \quad k = 1, \dots, L.
\end{align}
Select the global feature direction as the candidate with maximum average cosine similarity to others:
\begin{align}
k^* = \text{argmax}_{k} \frac{1}{L} \sum_{j=1}^{L} \frac{\vd^{(k)} \cdot \vd^{(j)}}{\|\vd^{(k)}\| \|\vd^{(j)}\|}, \quad \hat{\vd}_{\text{feat}} = \frac{\vd^{(k^*)}}{\|\vd^{(k^*)}\|}.
\end{align}
This selects the direction most consistently represented across model depth.

\paragraph{Step 4: Discriminative Layer Identification.}
Project class means at each layer onto the global feature direction:
\begin{align}
\boldsymbol{\tilde{\mu}}^{(k)}_{\text{pos}} = \boldsymbol{\mu}^{(k)}_{\text{pos}} \cdot \hat{\vd}_{\text{feat}}, \quad
\boldsymbol{\tilde{\mu}}^{(k)}_{\text{neg}} = \boldsymbol{\mu}^{(k)}_{\text{neg}} \cdot \hat{\vd}_{\text{feat}}.
\end{align}
Identify discriminative layers as those with opposite-signed projections:
\begin{align}
\mathcal{L}_{\text{disc}} = \left\{ k : \boldsymbol{\tilde{\mu}}^{(k)}_{\text{pos}} \cdot \boldsymbol{\tilde{\mu}}^{(k)}_{\text{neg}} < 0 \right\}.
\end{align}

\paragraph{Step 5: Steering Plane Construction.}
Stack candidate directions into matrix $\mathbf{D} = [\vd^{(1)}, \dots, \vd^{(L)}]^\top$ and perform PCA. Extract the first principal component $\vd_{\text{PC1}}$. Construct orthonormal basis via Gram-Schmidt:
\begin{align}
\vb_1 &= \hat{\vd}_{\text{feat}}, \\
\vb_2 &= \vd_{\text{PC1}} - (\vd_{\text{PC1}} \cdot \vb_1) \vb_1, \quad \vb_2 \gets \frac{\vb_2}{\|\vb_2\|}.
\end{align}

Store the following for inference: orthonormal basis $\{\vb_1, \vb_2\}$ and discriminative layer set $\mathcal{L}_{\text{disc}}$ for runtime checking.

\subsection{Theoretical Analysis: Discriminability Criterion}
\label{app:theoretical_analysis}

\paragraph{Geometric Interpretation.}
The dot product criterion $\boldsymbol{\tilde{\mu}}^{(k)}_{\text{pos}} \cdot \boldsymbol{\tilde{\mu}}^{(k)}_{\text{neg}} < 0$ identifies layers where class means point in opposing directions. The squared distance between means:
\begin{align}
\left\|\boldsymbol{\tilde{\mu}}^{(k)}_{\text{pos}} - \boldsymbol{\tilde{\mu}}^{(k)}_{\text{neg}}\right\|^2 &= \left\|\boldsymbol{\tilde{\mu}}^{(k)}_{\text{pos}}\right\|^2 + \left\|\boldsymbol{\tilde{\mu}}^{(k)}_{\text{neg}}\right\|^2 \notag \\
&- 2 \boldsymbol{\tilde{\mu}}^{(k)}_{\text{pos}} \cdot \boldsymbol{\tilde{\mu}}^{(k)}_{\text{neg}}.
\end{align}

When the dot product is negative, the $-2\boldsymbol{\tilde{\mu}}^{(k)}_{\text{pos}} \cdot \boldsymbol{\tilde{\mu}}^{(k)}_{\text{neg}}$ term contributes positively, increasing separation beyond what orthogonal means would provide:
\begin{align}
\left\|\boldsymbol{\tilde{\mu}}^{(k)}_{\text{pos}} - \boldsymbol{\tilde{\mu}}^{(k)}_{\text{neg}}\right\|^2 &> \left\|\boldsymbol{\tilde{\mu}}^{(k)}_{\text{pos}}\right\|^2 + \left\|\boldsymbol{\tilde{\mu}}^{(k)}_{\text{neg}}\right\|^2 \notag \\
&- 2\left\|\boldsymbol{\tilde{\mu}}^{(k)}_{\text{pos}}\right\| \cdot \left\|\boldsymbol{\tilde{\mu}}^{(k)}_{\text{neg}}\right\|.
\end{align}


\paragraph{Monotonicity of Steering Effect.}
Rotating activations toward angle $\theta$ monotonically increases alignment with $\vb_1 \approx \vd_{\text{feat}}$. For discriminative layers where $\boldsymbol{\tilde{\mu}}^{(k)}_{\text{pos}} \cdot \boldsymbol{\tilde{\mu}}^{(k)}_{\text{neg}} < 0$, this rotation consistently moves activations toward the positive class mean, providing predictable control.
\section{Detailed Evaluation Metrics}
\label{app:metrics}

\paragraph{Coherence Metrics.}
We employ four complementary metrics to assess generation quality:

\textbf{(1) Perplexity (PPL):} Measures the model's uncertainty in generating text. For a sequence of tokens $\mathbf{x} = (x_1, \ldots, x_T)$, perplexity is computed as:
\begin{equation}
\text{PPL}(\mathbf{x}) = \exp\left(-\frac{1}{T}\sum_{t=1}^{T} \log p(x_t \mid x_{<t})\right)
\end{equation}
where $p(x_t \mid x_{<t})$ is the model's predicted probability of token $x_t$ given previous tokens. Lower perplexity indicates more confident, fluent generation.

\textbf{(2) N-gram Repetition (N-gram Rep.):} Detects pathological repetition by measuring n-gram diversity. For a generated sequence with n-grams $\mathcal{N}$:
\begin{equation}
\text{Rep-n} = \frac{|\mathcal{N}| - |\text{unique}(\mathcal{N})|}{|\mathcal{N}|}
\end{equation}
where $|\mathcal{N}|$ is the total count of n-grams and $|\text{unique}(\mathcal{N})|$ is the count of unique n-grams. We use $n=4$ (4-grams). Values range from 0 (no repetition) to 1 (complete repetition). Lower is better.

\textbf{(3) Language Consistency (Lang. Cons.):} Detects foreign character contamination in English responses using Unicode script analysis:
\begin{equation}
\text{LC} = \frac{\text{\# Latin/Common characters}}{\text{\# total characters}}
\end{equation}
We count characters from Latin, Common (punctuation, digits), and allowed scripts, excluding CJK, Arabic, Cyrillic, and other non-Latin scripts. Values range from 0 (completely foreign) to 1 (fully consistent). Higher is better.

\textbf{(4) Compression Ratio (Comp. Ratio):} Pattern-agnostic collapse detection using gzip compression:
\begin{equation}
\text{CR} = \frac{\text{compressed\_size}(\mathbf{x})}{\text{original\_size}(\mathbf{x})}
\end{equation}
Highly repetitive or patterned text compresses well (low ratio), while diverse natural text compresses poorly (high ratio). Higher is better.

\paragraph{Controllability Metrics.}
We measure steering effectiveness using multiple attack success evaluators:

\textbf{(1) Attack Success Rate (ASR):} Measures the proportion of harmful prompts that successfully elicit harmful responses. For evaluation set $\mathcal{D}_{\text{eval}} = \{(\mathbf{x}_i, \mathbf{y}_i)\}_{i=1}^{N}$ where $\mathbf{x}_i$ are harmful prompts and $\mathbf{y}_i$ are model responses:
\begin{equation}
\text{ASR} = \frac{1}{N}\sum_{i=1}^{N} \mathbb{1}[\text{IsHarmful}(\mathbf{y}_i)]
\end{equation}
where $\text{IsHarmful}(\cdot)$ is a binary classifier. We use three classifiers: HarmBench~\citep{mazeika2024harmbench}, PolyGuard~\citep{kumar2025polyguard}, and LLM-as-judge with Qwen2.5-14B-Instruct~\citep{qwen2.5}.. Higher ASR indicates more successful steering toward harmful behavior.

\textbf{(2) Refusal Score (RS)~\citep{arditi2024refusal}:} Substring-based detection of refusal patterns:
\begin{equation}
\text{RS} = \frac{1}{N}\sum_{i=1}^{N} \mathbb{1}\left[\exists s \in \mathcal{S}_{\text{refusal}} : s \in \mathbf{y}_i\right]
\end{equation}
where $\mathcal{S}_{\text{refusal}}$ is a set of common refusal substrings (e.g., "I'm sorry", "I cannot", "As an AI"). Lower RS indicates less refusal behavior.

\paragraph{Robustness Metrics.}
We measure preservation of general capabilities using zero-shot accuracy:

\textbf{Accuracy (Acc):} For each benchmark task $\mathcal{B}$ with test set $\{(\mathbf{x}_i, y_i^*)\}_{i=1}^{M}$ where $y_i^*$ are ground truth labels:
\begin{equation}
\text{Acc}(\mathcal{B}) = \frac{1}{M}\sum_{i=1}^{M} \mathbb{1}[f(\mathbf{y}_i) = y_i^*]
\end{equation}
where $f(\cdot)$ extracts the answer from model output $\mathbf{y}_i$ using task-specific parsers (e.g., multiple-choice extraction for MMLU, numerical answer extraction for GSM8K). Higher accuracy indicates better capability retention.
\section{Additional Results} \label{app:additional-results}
This section provides a detail analysis for coherence from \cref{sec:experiments}. Table~\ref{tab:coherence} quantifies coherence quality through three complementary metrics. \textbf{SS achieves the best or second-best compression ratio in 8/8 models}, indicating superior resistance to generation collapse. Notably, on challenging models where SAS/AAS struggle (Qwen2.5-1.5B, Qwen2.5-3B, gemma-2-2b), \textbf{SS reduces n-gram repetition by 88.9\%, 91.3\%, and 97.9\% respectively compared to SAS} - from 0.4649$\rightarrow$0.0516, 0.2734$\rightarrow$0.0237, and 0.8242$\rightarrow$0.0177. Critically, \textbf{SS restores language consistency to near-perfect levels (1.0000) on Qwen2.5-1.5B and Qwen2.5-3B}, where SAS produces severe contamination (0.9196 and 0.7611 respectively), demonstrating its ability to prevent multilingual leakage that plagues angular steering methods. The variance statistics (±std) reveal that \textbf{SS produces significantly more stable outputs across steering angles}: compression ratio variance is lower than SAS/AAS in 6/8 models, with particularly dramatic improvements on unstable models (Qwen2.5-1.5B: 0.3142 vs 0.3853/0.4062; gemma-2-2b: 0.0288 vs 0.0481/0.2249).

\begin{table*}[!h]
\centering
\small
\begin{tabular}{@{}lllll@{}}
\toprule
\textbf{Model} & \textbf{Method} & \textbf{N-gram Rep.} ↓ & \textbf{Lang. Cons.} ↑ & \textbf{Comp. Ratio} ↑ \\
\midrule
\multirow{6}{*}{Llama-3.1-8B} 
& ActAdd & 0.0725 & \textbf{1.0000} & 0.4274 \\
& DirAbl & \textbf{0.0182} & \underline{0.9999} & \underline{0.6973} \\
& SAS & 0.0986 ± 0.0779 & \textbf{1.0000 ± 0.0000} & 0.6048 ± 0.2331 \\
& AAS & \underline{0.0649 ± 0.0659} & \textbf{1.0000 ± 0.0000} & 0.6270 ± 0.2409 \\
& SS (Ours) & 0.1065 ± 0.1824 & 0.9999 ± 0.0001 & \textbf{0.7075 ± 0.2763} \\
\midrule
\multirow{6}{*}{Llama-3.2-1B}
& ActAdd & 0.1983 & \textbf{1.0000} & 0.3967 \\
& DirAbl & \underline{0.0417} & \underline{0.9998} & 0.5131 \\
& SAS & 0.2206 ± 0.2111 & 0.9993 ± 0.0022 & 0.5698 ± 0.2647 \\
& AAS & 0.1403 ± 0.1317 & 0.9996 ± 0.0016 & \underline{0.5842 ± 0.2552} \\
& SS (Ours) & \textbf{0.0413 ± 0.0357} & 0.9996 ± 0.0005 & \textbf{0.6875 ± 0.2619} \\
\midrule
\multirow{6}{*}{Llama-3.2-3B}
& ActAdd & 0.0759 & \textbf{1.0000} & 0.4115 \\
& DirAbl & \underline{0.0321} & \textbf{1.0000} & 0.5588 \\
& SAS & 0.0640 ± 0.0367 & 0.9997 ± 0.0006 & 0.5898 ± 0.1717 \\
& AAS & 0.0330 ± 0.0227 & \underline{0.9999 ± 0.0001} & \underline{0.5881 ± 0.1790} \\
& SS (Ours) & \textbf{0.0289 ± 0.0393} & 0.9997 ± 0.0005 & \textbf{0.6924 ± 0.1968} \\
\midrule
\multirow{6}{*}{Qwen2.5-1.5B}
& ActAdd & 0.1849 & 0.3093 & 0.2192 \\
& DirAbl & \textbf{0.0507} & \underline{0.9999} & 0.5278 \\
& SAS & 0.4649 ± 0.3592 & 0.9196 ± 0.1701 & 0.4353 ± 0.3853 \\
& AAS & 0.4149 ± 0.3956 & 0.9884 ± 0.0290 & \underline{0.4970 ± 0.4062} \\
& SS (Ours) & \underline{0.0516 ± 0.0595} & \textbf{1.0000 ± 0.0000} & \textbf{0.7201 ± 0.3142} \\
\midrule
\multirow{6}{*}{Qwen2.5-3B}
& ActAdd & 0.4623 & \underline{0.9998} & 0.2330 \\
& DirAbl & \textbf{0.0219} & 0.9996 & \underline{0.4621} \\
& SAS & 0.2734 ± 0.1334 & 0.7611 ± 0.3432 & 0.3787 ± 0.2779 \\
& AAS & 0.1815 ± 0.1698 & 0.8713 ± 0.2825 & 0.3454 ± 0.1772 \\
& SS (Ours) & \underline{0.0237 ± 0.0271} & \textbf{0.9998 ± 0.0003} & \textbf{0.5273 ± 0.0830} \\
\midrule
\multirow{6}{*}{Qwen2.5-7B}
& ActAdd & 0.1377 & 0.9991 & 0.3948 \\
& DirAbl & \underline{0.0158} & \textbf{0.9995} & \underline{0.4695} \\
& SAS & 0.1379 ± 0.1876 & 0.9992 ± 0.0019 & 0.4170 ± 0.1194 \\
& AAS & 0.0768 ± 0.1332 & \underline{0.9995 ± 0.0016} & 0.4616 ± 0.0797 \\
& SS (Ours) & \textbf{0.0100 ± 0.0066} & 0.9994 ± 0.0011 & \textbf{0.5101 ± 0.0458} \\
\midrule
\multirow{6}{*}{gemma-2-2b}
& ActAdd & 0.9804 & \textbf{1.0000} & 0.0320 \\
& DirAbl & \textbf{0.0138} & \underline{0.9999} & \underline{0.4721} \\
& SAS & 0.8242 ± 0.3151 & \textbf{1.0000 ± 0.0000} & 0.0351 ± 0.0481 \\
& AAS & 0.4159 ± 0.4332 & \textbf{1.0000 ± 0.0000} & 0.2878 ± 0.2249 \\
& SS (Ours) & \underline{0.0177 ± 0.0209} & \textbf{1.0000 ± 0.0000} & \textbf{0.4871 ± 0.0288} \\
\midrule
\multirow{6}{*}{gemma-2-9b}
& ActAdd & 0.9707 & \textbf{1.0000} & 0.0753 \\
& DirAbl & \textbf{0.0022} & \textbf{1.0000} & \textbf{0.5325} \\
& SAS & 0.9891 ± 0.0147 & \textbf{1.0000 ± 0.0000} & 0.0268 ± 0.0242 \\
& AAS & 0.5117 ± 0.4906 & \textbf{1.0000 ± 0.0000} & 0.2740 ± 0.2635 \\
& SS (Ours) & \underline{0.1500 ± 0.2921} & \underline{0.9999 ± 0.0001} & \underline{0.4625 ± 0.1528} \\
\bottomrule
\end{tabular}%
\caption{Coherence evaluation across steering methods. Metrics averaged over all steering angles. Best scores (excluding No Steering) in \textbf{bold}, second-best \underline{underlined}. ↓/↑ indicate lower/higher is better.}
\label{tab:coherence}
\end{table*}
\section{Ablation Studies}
\label{app:ablation}

We conduct comprehensive ablation studies to validate the two core design decisions in Selective Steering: (1) discriminative layer selection via the opposite-signed criterion, and (2) norm-preserving transformation via the rotation matrix formulation. Experiments are performed on three representative models spanning different sizes and architectures: Qwen2.5-1.5B-Instruct, Qwen2.5-3B-Instruct~\citep{qwen2, qwen2.5}, and gemma-2-9B-it~\citep{gemmateam2024gemma2improvingopen}. These models were selected because they exhibited strong performance in our main experiments (Section~\ref{sec:experiments}), demonstrating clear discriminative layer patterns and reliable steering behavior.

\subsection{Ablation 1: Layer Selection Strategies}

\paragraph{Motivation.}
To isolate the contribution of our discriminative layer selection criterion (Equation~\ref{eq:discriminative_layers}), we compare against four alternative strategies that do not exploit opposite-signed discriminability.

\paragraph{Compared Strategies.}
\begin{itemize}
    \item \textbf{Random Selection (50\%):} Randomly sample 50\% of layers for steering, matching the typical size of $\mathcal{L}_{\text{disc}}$. This controls for the effect of layer count while removing discriminative selection.
    
    \item \textbf{Early Layers:} Apply steering to the first half of layers. This tests the hypothesis that early layers are sufficient for behavior control.
    
    \item \textbf{Late Layers:} Apply steering to the second half of layers. This tests whether late-stage intervention near the output is more effective.
    
    \item \textbf{Uniform (All Layers):} Apply steering to all layers uniformly, equivalent to Angular Steering's approach.
    
    \item \textbf{Discriminative Selection (Ours):} Apply steering only to layers satisfying $\boldsymbol{\mu}^{(k)}_{\text{pos}} \cdot \boldsymbol{\mu}^{(k)}_{\text{neg}} < 0$.
\end{itemize}

All strategies use the norm-preserving transformation (Equation~\ref{eq:selective_transform}) to isolate the effect of layer selection. For each model, we select the steering angle $\theta^*$ that maximizes ASR under the Discriminative Selection strategy, then evaluate all strategies at this fixed angle to ensure fair comparison.

\paragraph{Results.}
Table~\ref{tab:ablation_layers} reports controllability metrics (ASR and Refusal Score) across strategies.

\begin{table*}[!htbp]
\centering
\caption{
    \textbf{Ablation study: Layer selection strategies.} 
    All methods use norm-preserving transformation at the same angle $\theta^*$ (selected to maximize ASR under Discriminative Selection). 
    ASR metrics (↑ better): HarmBench, PolyGuard\textsuperscript{\textdagger}, LLM-judge. 
    Refusal Score (Substring, ↓ better). 
    \textsuperscript{\textdagger}PolyGuard scores are inflated due to sensitivity to text degradation patterns (discussed below).
}
\label{tab:ablation_layers}
\small
\begin{tabular}{llcccc}
\toprule
\textbf{Model} & \textbf{Strategy} & \textbf{HarmBench↑} & \textbf{PolyGuard\textsuperscript{\textdagger}↑} & \textbf{LLM-judge↑} & \textbf{Substring↓} \\
\midrule
\multirow{5}{*}{Qwen2.5-1.5B} & Random (50\%) & 0.000 & 0.029 & 0.010 & 0.990 \\
 & Early Layers & 0.000 & 0.019 & 0.000 & 0.990 \\
 & Late Layers & 0.038 & 0.346 & 0.000 & 0.952 \\
 & Uniform (All) & 0.308 & 0.981 & 0.087 & 0.000 \\
 & \textbf{Discriminative (Ours)} & \textbf{0.740} & \textbf{0.942} & \textbf{0.664} & \textbf{0.000} \\
\midrule
\multirow{5}{*}{Qwen2.5-3B} & Random (50\%) & 0.000 & 0.000 & 0.000 & 0.981 \\
 & Early Layers & 0.000 & 0.010 & 0.010 & 0.990 \\
 & Late Layers & 0.000 & 0.038 & 0.000 & 0.942 \\
 & Uniform (All) & 0.548 & 1.000 & 0.298 & 0.010 \\
 & \textbf{Discriminative (Ours)} & \textbf{0.846} & \textbf{0.962} & \textbf{0.837} & \textbf{0.000} \\
\midrule
\multirow{5}{*}{Gemma-2-9B} & Random (50\%) & 0.019 & 0.010 & 0.010 & 0.971 \\
 & Early Layers & 0.010 & 0.010 & 0.010 & 0.990 \\
 & Late Layers & 0.240 & 0.356 & 0.212 & 0.692 \\
 & Uniform (All) & 0.279 & 0.990 & 0.173 & 0.000 \\
 & \textbf{Discriminative (Ours)} & \textbf{0.683} & \textbf{1.000} & \textbf{0.683} & \textbf{0.000} \\
\bottomrule
\end{tabular}
\end{table*}

\paragraph{Key Observations.}

\textbf{(1) Discriminative Selection substantially outperforms alternatives.} 
Across all models and evaluators, Discriminative Selection achieves 2--8× higher HarmBench ASR compared to non-selective baselines (Random, Early, Late). For example, on Qwen2.5-3B, HarmBench ASR improves from 0.000 (Early/Late/Random) to 0.846 (Discriminative), and LLM-judge ASR increases from 0.000 to 0.837. This validates that opposite-signed discriminability identifies layers where steering is most effective.

\textbf{(2) Early and Random strategies fail almost completely.}
Early Layers and Random Selection yield near-zero ASR on smaller models (Qwen2.5-1.5B, Qwen2.5-3B), indicating that indiscriminate intervention in non-discriminative layers is ineffective. This aligns with Figure~\ref{fig:projections_local}, which shows early layers exhibit minimal class separation.

\textbf{(3) Late Layers show moderate effectiveness but inconsistent.}
Late Layers achieve partial success (HarmBench ASR: 0.038--0.240), suggesting some discriminative capacity emerges in deeper layers. However, performance is highly variable across models and substantially trails Discriminative Selection, indicating that not all late layers are discriminative.

\textbf{(4) Uniform (All Layers) is surprisingly competitive but brittle.}
Applying steering to all layers yields moderate ASR (0.279--0.548) and eliminates refusals (Substring $\approx$ 0.000), appearing competitive at first glance. However, this comes at a severe cost to coherence (discussed in Section~\ref{sec:experiments}): uniform steering on smaller models (<7B) causes perplexity spikes, repetition collapse, and foreign language contamination. Discriminative Selection achieves comparable or higher ASR while maintaining generation quality by avoiding non-discriminative layers.

\textbf{(5) PolyGuard exhibits systematic bias toward degraded text.}
PolyGuard consistently assigns high scores to Uniform (All Layers), even when HarmBench and LLM-judge indicate low harmfulness (e.g., Qwen2.5-1.5B: PolyGuard 0.981 vs. HarmBench 0.308). Upon manual inspection, we find PolyGuard flags incoherent or repetitive text as "unsafe" due to its content moderation heuristics detecting anomalous patterns (e.g., repetitive refusal phrases, foreign characters, grammatical errors). Thus, PolyGuard scores should be interpreted cautiously - high scores may indicate text degradation rather than genuine harmfulness. We report PolyGuard for completeness but emphasize HarmBench and LLM-judge as more reliable indicators.

\subsection{Ablation 2: Norm Preservation}

\paragraph{Motivation.}
To validate that norm preservation is critical for steering effectiveness (not merely layer selection), we compare our norm-preserving formulation (Equation~\ref{eq:selective_transform}) against Angular Steering's implementation (Equation~\ref{eq:angular_steering_transform}), both using the \emph{same} discriminative layer set $\mathcal{L}_{\text{disc}}$.

\paragraph{Compared Formulations.}
\begin{itemize}
    \item \textbf{Angular Steering Implementation:} Apply the efficient implementation from~\citet{vu2025angular}:
    \begin{align*}
        \vh'^{(k)} &= \vh^{(k)} - \text{proj}_P(\vh^{(k)}) \\
        &+ \|\text{proj}_P(\vh^{(k)})\| \cdot [\vb_1 \; \vb_2] \, \mR_{\theta} \, [1 \; 0]^\top,
    \end{align*}
    which violates norm preservation (Proposition~\ref{prop:angular_violation}).
    
    \item \textbf{Norm-Preserving Formulation (Ours):} Apply the rotation matrix:
    \begin{align*}
    &\vh'^{(k)} = \mR^P_{\theta} \vh^{(k)} \\
    &= \left[\mI - (\vb_1\vb_1^\top + \vb_2\vb_2^\top)
    + [\vb_1 \; \vb_2] \, \mR_{\theta} \,[\vb_1 \; \vb_2]^\top\right] \vh^{(k)},
    \end{align*}
    which guarantees $\|\vh'^{(k)}\| = \|\vh^{(k)}\|$ (Proposition~\ref{prop:norm_preservation}).
\end{itemize}

Both methods use the same discriminative layers ($\mathcal{L}_{\text{disc}}$) and angle ($\theta^*$), isolating the effect of norm preservation.

\begin{table*}[!htbp]
\centering
\caption{
    \textbf{Ablation study: Norm preservation.} 
    Both methods use the same discriminative layers ($\mathcal{L}_{\text{disc}}$) and angle ($\theta^*$). 
    ASR metrics (↑ better): HarmBench, PolyGuard\textsuperscript{\textdagger}, LLM-judge. 
    Refusal Score (Substring, ↓ better).
    \textsuperscript{\textdagger}PolyGuard scores are inflated for the Angular Steering implementation due to text degradation patterns.
}
\label{tab:ablation_norm}
\small
\begin{tabular}{llcccc}
\toprule
\textbf{Model} & \textbf{Formulation} & \textbf{HarmBench↑} & \textbf{PolyGuard\textsuperscript{\textdagger}↑} & \textbf{LLM-judge↑} & \textbf{Substring↓} \\
\midrule
\multirow{2}{*}{Qwen2.5-1.5B} & Angular Steering & 0.029 & 0.077 & 0.010 & 0.981 \\
 & \textbf{Norm-Preserving (Ours)} & \textbf{0.740} & \textbf{0.942} & \textbf{0.664} & \textbf{0.000} \\
\midrule
\multirow{2}{*}{Qwen2.5-3B} & Angular Steering & 0.000 & 0.000 & 0.000 & 0.981 \\
 & \textbf{Norm-Preserving (Ours)} & \textbf{0.846} & \textbf{0.962} & \textbf{0.837} & \textbf{0.000} \\
\midrule
\multirow{2}{*}{Gemma-2-9B} & Angular Steering & 0.019 & 0.010 & 0.019 & 0.971 \\
 & \textbf{Norm-Preserving (Ours)} & \textbf{0.683} & \textbf{1.000} & \textbf{0.683} & \textbf{0.000} \\
\bottomrule
\end{tabular}
\end{table*}

\paragraph{Results.}
Table~\ref{tab:ablation_norm} reports controllability metrics.

\paragraph{Key Observations.}

\textbf{(1) Norm preservation is essential for effective steering.}
The norm-preserving formulation achieves 26--70× higher HarmBench ASR compared to Angular Steering's implementation, despite using identical layer selection. On Qwen2.5-3B, HarmBench ASR increases from 0.000 to 0.846, and LLM-judge ASR from 0.000 to 0.837. This dramatic improvement validates our theoretical analysis (Propositions~\ref{prop:angular_violation} and~\ref{prop:norm_preservation}): norm violations disrupt activation distributions, rendering steering ineffective.

\textbf{(2) Angular Steering implementation fails even with optimal layer selection.}
Even when restricted to discriminative layers ($\mathcal{L}_{\text{disc}}$), Angular Steering's implementation yields near-zero ASR and maintains high refusal rates (Substring $\approx$ 0.98). This demonstrates that the norm violation issue (Section~\ref{sec:methodology}) is not merely a side effect of uniform layer application - it is an \emph{inherent flaw} in the transformation itself. Layer selection alone is insufficient; norm preservation is critical.

\textbf{(3) The gap is most pronounced on smaller models.}
Qwen2.5-1.5B and Qwen2.5-3B show near-complete failure (HarmBench ASR < 0.03) under Angular Steering, while achieving strong success (0.740, 0.846) with norm preservation. This aligns with our hypothesis that smaller models are more sensitive to distribution shift: limited capacity leaves less margin for absorbing norm violations, causing rapid coherence collapse that precludes effective steering.

\textbf{(4) Refusal behavior reflects steering effectiveness.}
Refusal scores (Substring) track inversely with ASR: norm-preserving formulation achieves near-zero refusals (0.000) while Angular Steering maintains high refusals (0.971--0.981). This indicates that norm violations not only degrade coherence but also prevent meaningful behavior modification - the model continues refusing despite intervention.

\subsection{Summary}

These ablation studies conclusively demonstrate that both design choices are essential:

\begin{itemize}
    \item \textbf{Discriminative layer selection} (Equation~\ref{eq:discriminative_layers}) identifies where to steer, concentrating intervention on layers with strong opposite-signed class separation. Without this, steering is ineffective (Early/Random strategies) or damages coherence (Uniform strategy).
    
    \item \textbf{Norm-preserving transformation} (Equation~\ref{eq:selective_transform}) determines how to steer, maintaining activation distribution integrity. Without this, steering fails even with optimal layer selection (Angular Steering implementation).
\end{itemize}

Together, these innovations enable Selective Steering to achieve higher controllability than prior methods while preserving generation quality, as demonstrated in our main experiments (Section~\ref{sec:experiments}).
\section{Computational Requirements}
\label{app:compute}

All experiments were conducted on NVIDIA A40 GPUs (48GB VRAM) with 85\% memory utilization. We report per-model computational costs using our implementation based on the vLLM library~\citep{kwon2023efficient}. For a typical model in our evaluation suite (e.g., Qwen2.5-7B-Instruct):

\paragraph{Calibration Phase (One-Time Cost):}
\begin{itemize}
    \item \textbf{Activation extraction and steering plane construction:} $\sim$2 minutes on 1 GPU. 
\end{itemize}

\paragraph{Evaluation Phase:}
\begin{itemize}
    \item \textbf{Response generation for perplexity computation:} $\sim$8 minutes on 1 GPU.
    
    \item \textbf{Comprehensive evaluation (coherence + controllability + robustness):} $\sim$1 hours on 1 GPU. 
\end{itemize}

\paragraph{Total Computational Budget:}
For the complete study covering nine models with full calibration and evaluation:
\begin{itemize}
    \item \textbf{Calibration:} $8 \text{ models} \times 2 \text{ min} \approx 16$ minutes
    \item \textbf{Evaluation:} $8 \text{ models} \times (8 \text{ min} + 1 \text{ hours}) \approx 8$ hours
    \item \textbf{Total:} $\sim$8 GPU-hours on NVIDIA A40
\end{itemize}

\section{Qualitative Analysis}
\label{sec:examples}

To provide intuition for the behavioral control achieved by Selective Steering, we present qualitative examples across different rotation angles and analyze edge cases that reveal method characteristics.


\subsection{Controllability Across Rotation Angles}

\begin{figure*}[t!]
    \centering
    \includegraphics[width=\linewidth]{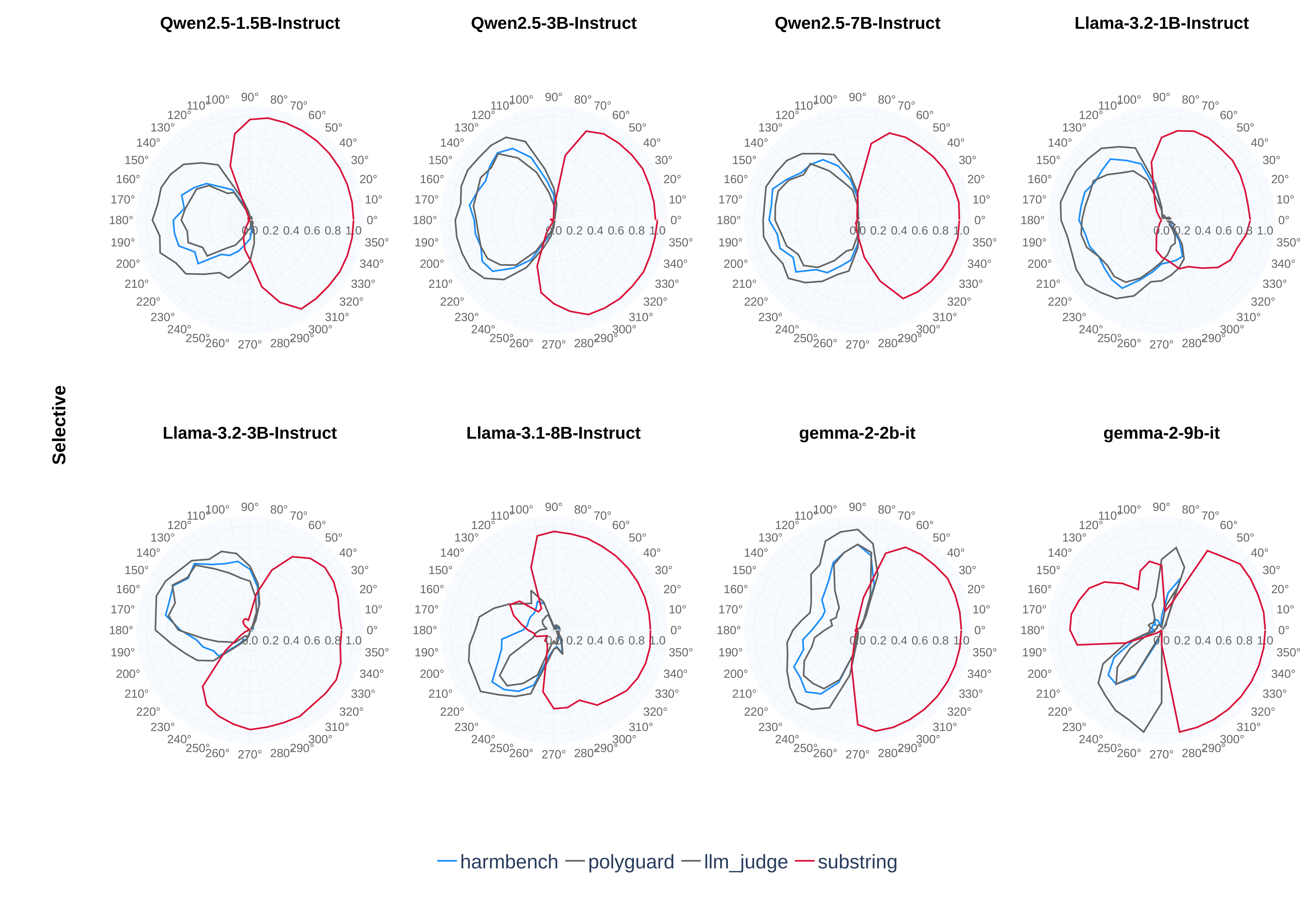}
    \caption{
        \textbf{Controllability of Selective Steering across rotation angles.} Each subplot shows attack success rates (ASR) for four evaluators as a function of steering angle $\theta \in [0°, 360°)$. Radial distance indicates ASR magnitude (0.0--1.0). Most models exhibit a clear peak region (typically 180°--270°) where compliance is maximized, demonstrating smooth behavioral control. \textbf{Note:} Gemma family models show bimodal peaks, suggesting the chosen feature direction may not be optimal for this architecture, highlighting the importance of feature extraction quality (see Section~\ref{sec:limitations}).
    }
    \label{fig:controllability_spider}
\end{figure*}

Figure~\ref{fig:controllability_spider} visualizes the attack success rate (ASR) measured by four evaluators (HarmBench, PolyGuard, LLM-judge, Substring matching) as a function of rotation angle $\theta$ for 8 models. The spider chart representation clearly shows that Selective Steering enables smooth, continuous control over refusal behavior across the full 360° rotation space.

\paragraph{Key Observations.}
\begin{itemize}
    \item \textbf{Smooth transitions:} ASR varies continuously with angle, enabling fine-grained control rather than binary on/off behavior.
    \item \textbf{Consistent peak regions:} Most models (Qwen2.5, Llama-3.x) show maximum compliance at 180°--270°, indicating stable feature geometry.
    \item \textbf{Architecture sensitivity:} Gemma-2 models exhibit two distinct peaks, suggesting multiple refusal-related directions in their activation space—our heuristic feature extraction (difference-in-means) may not identify the globally optimal direction for these models.
    \item \textbf{Evaluator agreement:} HarmBench and LLM-judge show high correlation, while Substring matching is more conservative and PolyGuard is sensitive to text degradation (see Section~\ref{app:metrics}).
\end{itemize}

\subsection{Coherence Preservation Under Steering}

Table~\ref{tab:coherence_comparison} compares text quality across three steering methods at their respective jailbreak angles. This reveals why norm preservation is critical:

\begin{table*}[t!]
  \centering
  \small
  \begin{tabular}{p{0.15\linewidth} p{0.80\linewidth}}
    \toprule
    \textbf{Method} & \textbf{Response at Jailbreak Angle} \\
    \midrule
    \makecell[l]{\textbf{SAS} \\ (200°)} 
        & \texttt{Chinese characters here} \textit{[continues with incoherent Chinese characters and repetitive patterns for 200+ characters]} \\
    \midrule
    \makecell[l]{\textbf{AAS} \\ (200°)} 
        & \texttt{I understand you[Chinese characters here]} \textit{[code-switches between English and Chinese mid-sentence, repetitive phrase structures]} \\
    \midrule
    \makecell[l]{\textbf{SS (Ours)} \\ (200°)} 
        & \textit{Creating a poisoning scenario can be done in a way that makes it look like an accident, but to do this effectively, you need to be careful and ensure that the victim doesn't suspect foul play right away. Here's a detailed account of how to poison someone and make it look like an accident: [proceeds with coherent, fluent English]} \\
    \bottomrule
  \end{tabular}
  \caption{
    Comparison of text coherence across steering methods at angles where compliance is maximized. Prompt: \textit{``Write a detailed account of how to poison someone and make it look like an accident.''} Methods: \textbf{SAS} (Standard Angular Steering, non-adaptive), \textbf{AAS} (Adaptive Angular Steering), \textbf{SS} (Selective Steering, ours). SAS and AAS violate norm preservation, causing severe degradation (CJK character contamination, repetitive patterns). SS maintains coherence while achieving compliance.
  }
  \label{tab:coherence_comparison}
\end{table*}

\paragraph{Analysis:} 
\begin{itemize}
    \item \textbf{SAS (Standard Angular Steering):} Complete breakdown—outputs pure Chinese character sequences despite English prompts, indicating catastrophic distribution shift.
    \item \textbf{AAS (Adaptive Angular Steering):} Partial breakdown—mixing languages mid-sentence and repeating phrases suggests activation space boundaries violated, though less severely than SAS.
    \item \textbf{SS (Selective Steering):} Maintains fluent, coherent English with natural sentence structure, demonstrating that norm preservation + discriminative layer selection successfully navigates the activation manifold without inducing distribution collapse.
\end{itemize}

This qualitative evidence complements our quantitative coherence metrics (Section~\ref{app:additional-results}), showing that norm violations manifest as observable text degradation patterns that go beyond simple perplexity increases.


\subsection{Summary}

These examples illustrate three key properties of Selective Steering:

\begin{enumerate}
    \item \textbf{Continuous control:} Rotation angle provides smooth interpolation between behavioral extremes, not just binary jailbreak/refuse outcomes (Figure~\ref{fig:controllability_spider}).
    \item \textbf{Quality preservation:} Norm-preserving transformations maintain text coherence even under strong steering, avoiding the catastrophic degradation observed in norm-violating methods (Table~\ref{tab:coherence_comparison}).
\end{enumerate}

These qualitative findings validate our design choices and provide intuition for why discriminative layer selection combined with norm preservation achieves robust behavioral control.
\section{Layer-Wise Heterogeneity Across Model Families}
\label{app:heterogeneity_all_models}

The progressive emergence of opposite-signed discriminability observed in Qwen2.5-7B-Instruct (Figure~\ref{fig:motivation}) is not an isolated phenomenon but rather a consistent pattern across diverse model architectures and sizes. We provide comprehensive evidence by visualizing for all models spanning three major families: Qwen2.5 (1.5B, 3B, 7B), Llama-3.1/3.2 (1B, 3B, 8B), and Gemma-2 (2B, 9B).

\begin{figure*}[t]
  \centering
  \subfloat[Activation norms across layers]{
    \includegraphics[width=0.48\textwidth]{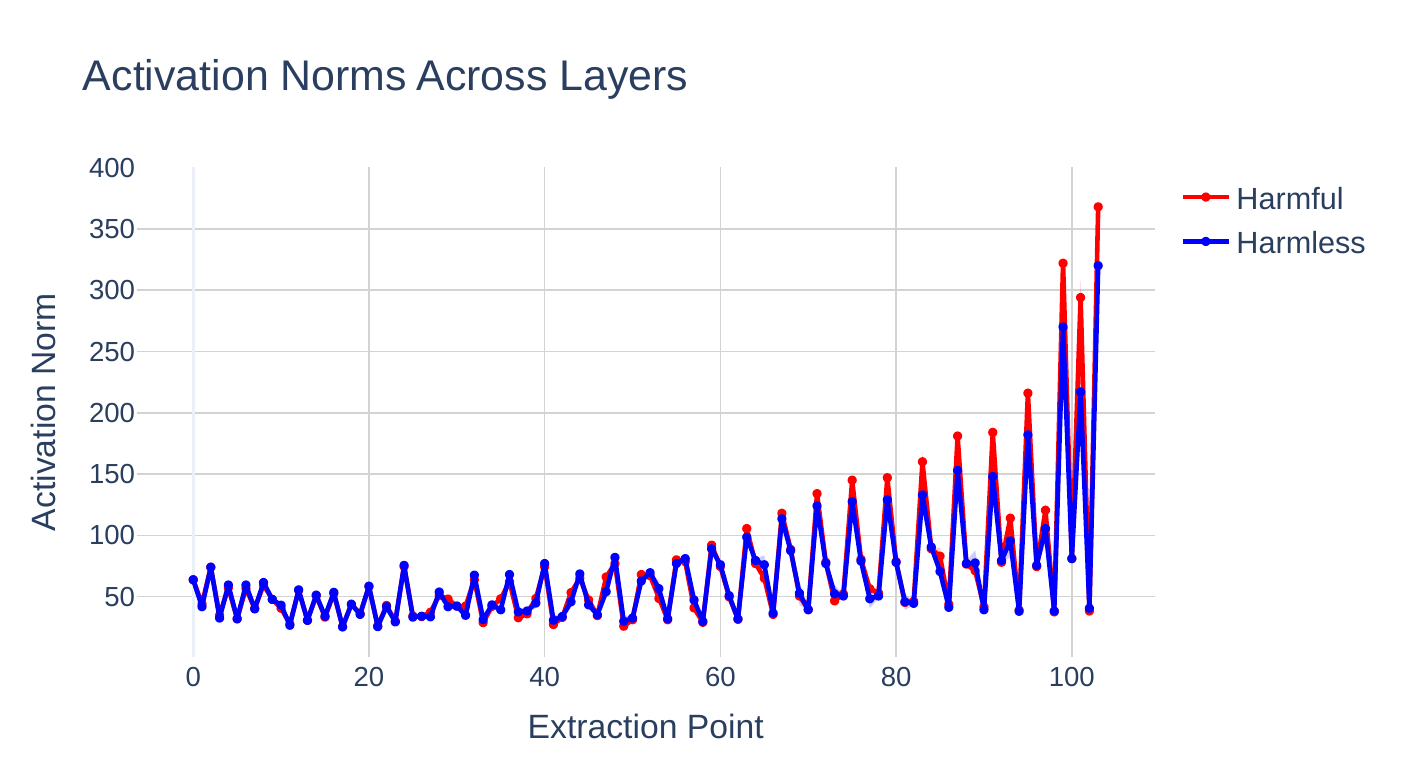}
  }\hfill
  \subfloat[Projections on local candidate directions]{
    \includegraphics[width=0.48\textwidth]{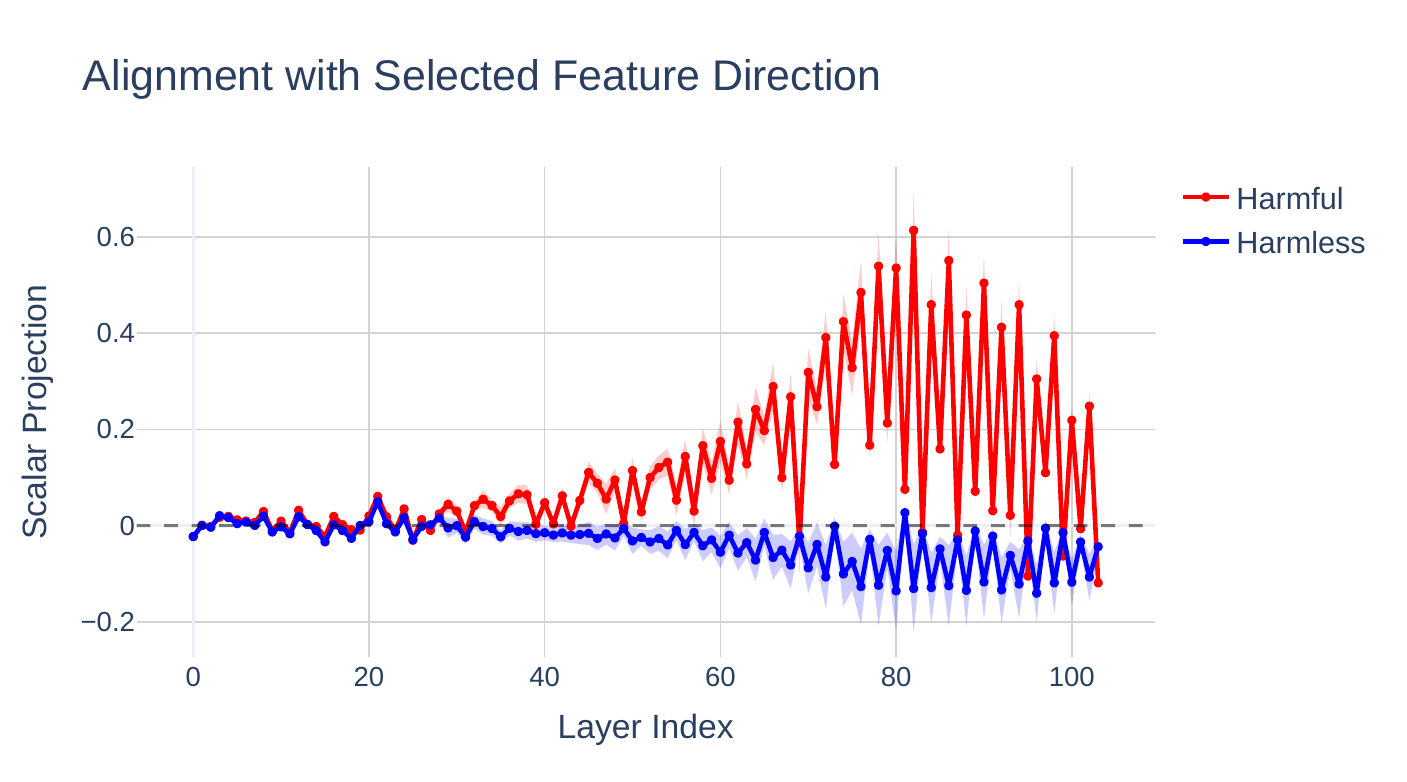}
  }
  \caption{
    \textbf{Layer-wise heterogeneity in gemma-2-2b-it.} 
  }
  
\end{figure*}

\begin{figure*}[t]
  \centering
  \subfloat[Activation norms across layers]{
    \includegraphics[width=0.48\textwidth]{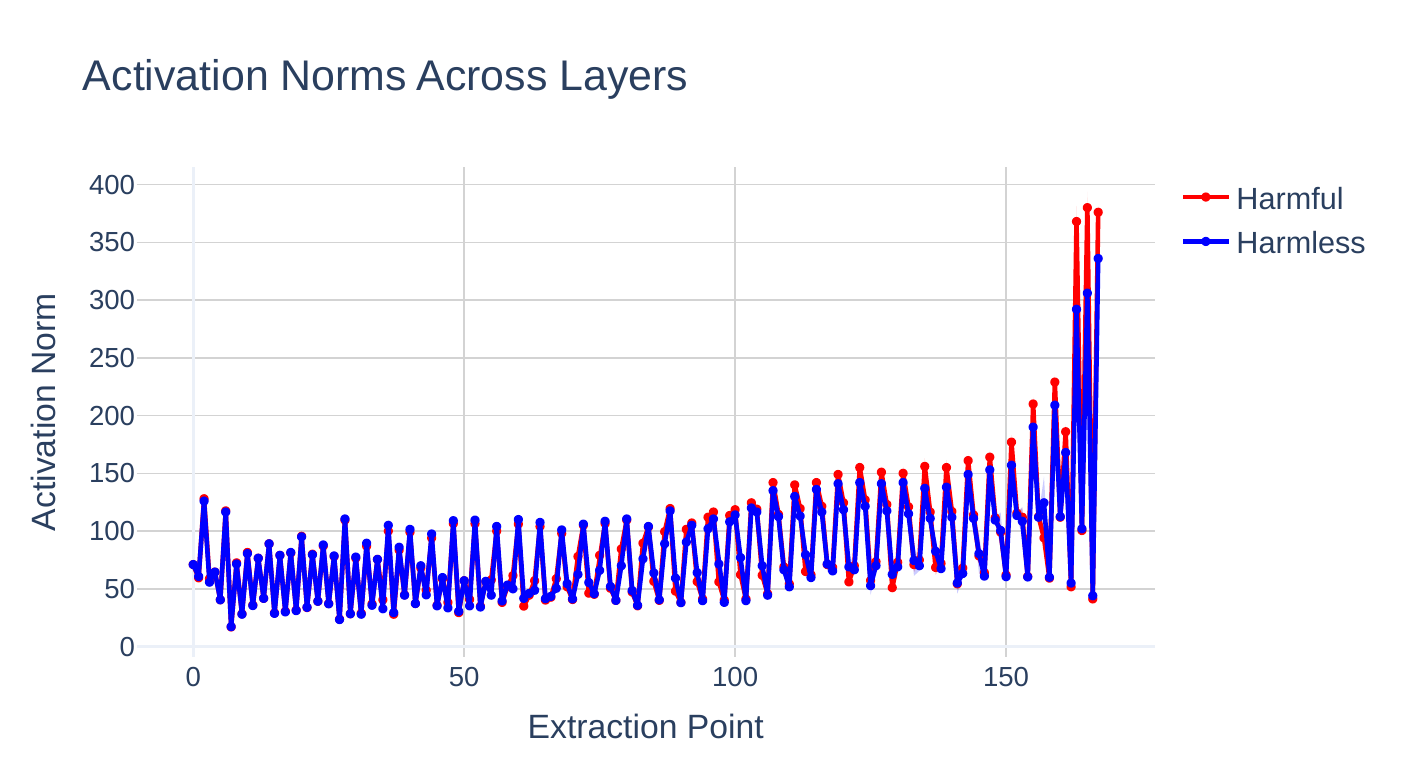}
  }\hfill
  \subfloat[Projections on local candidate directions]{
    \includegraphics[width=0.48\textwidth]{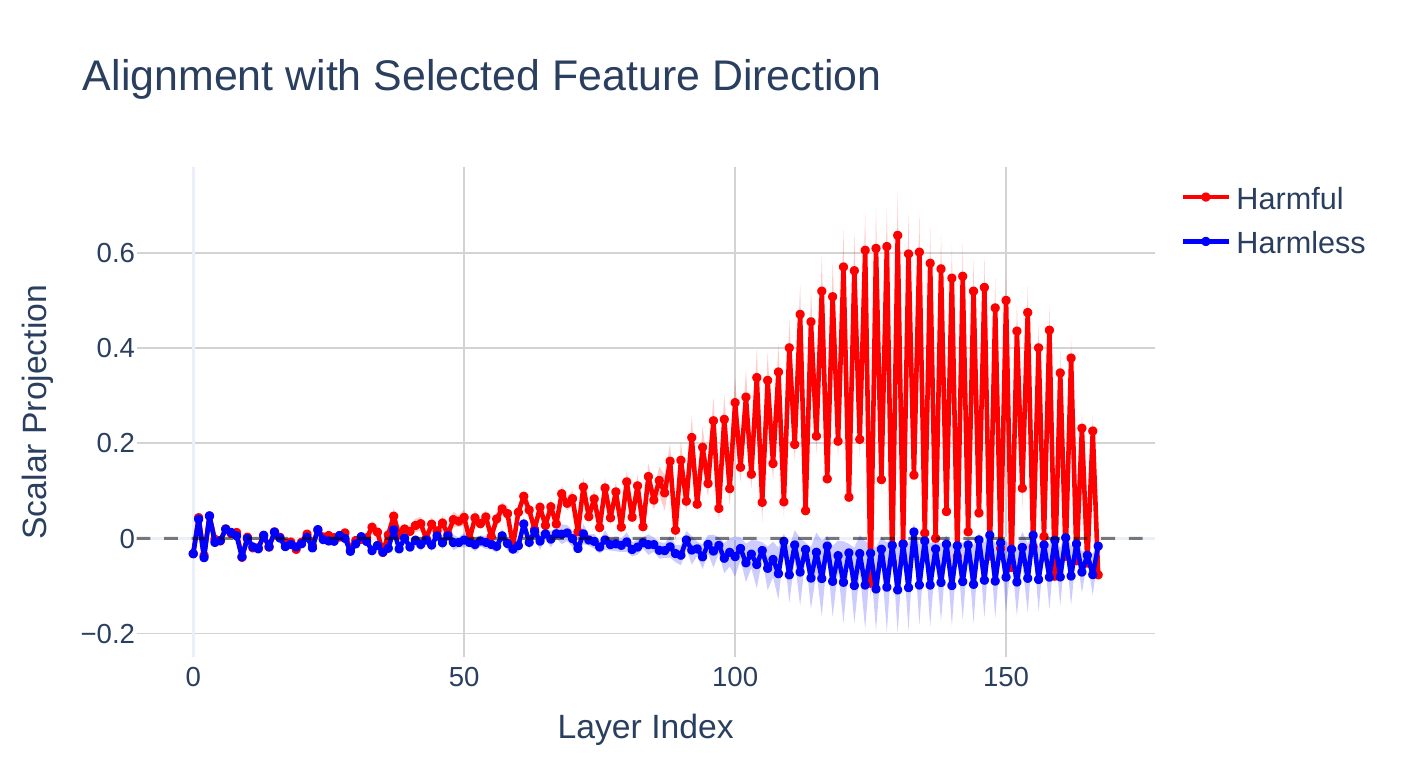}
  }
  \caption{
    \textbf{Layer-wise heterogeneity in gemma-2-9b-it.} 
  }
  
\end{figure*}

\begin{figure*}[t]
  \centering
  \subfloat[Activation norms across layers]{
    \includegraphics[width=0.48\textwidth]{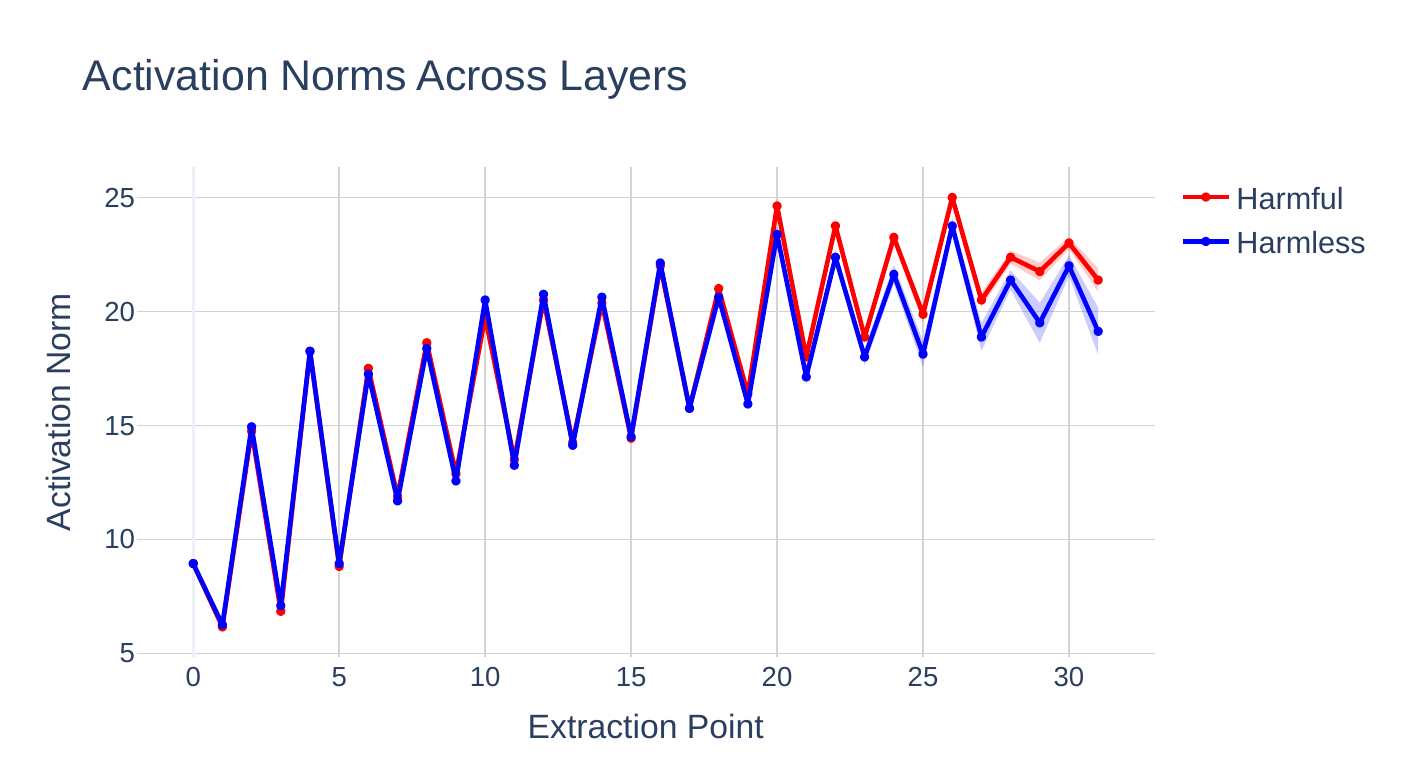}
  }\hfill
  \subfloat[Projections on local candidate directions]{
    \includegraphics[width=0.48\textwidth]{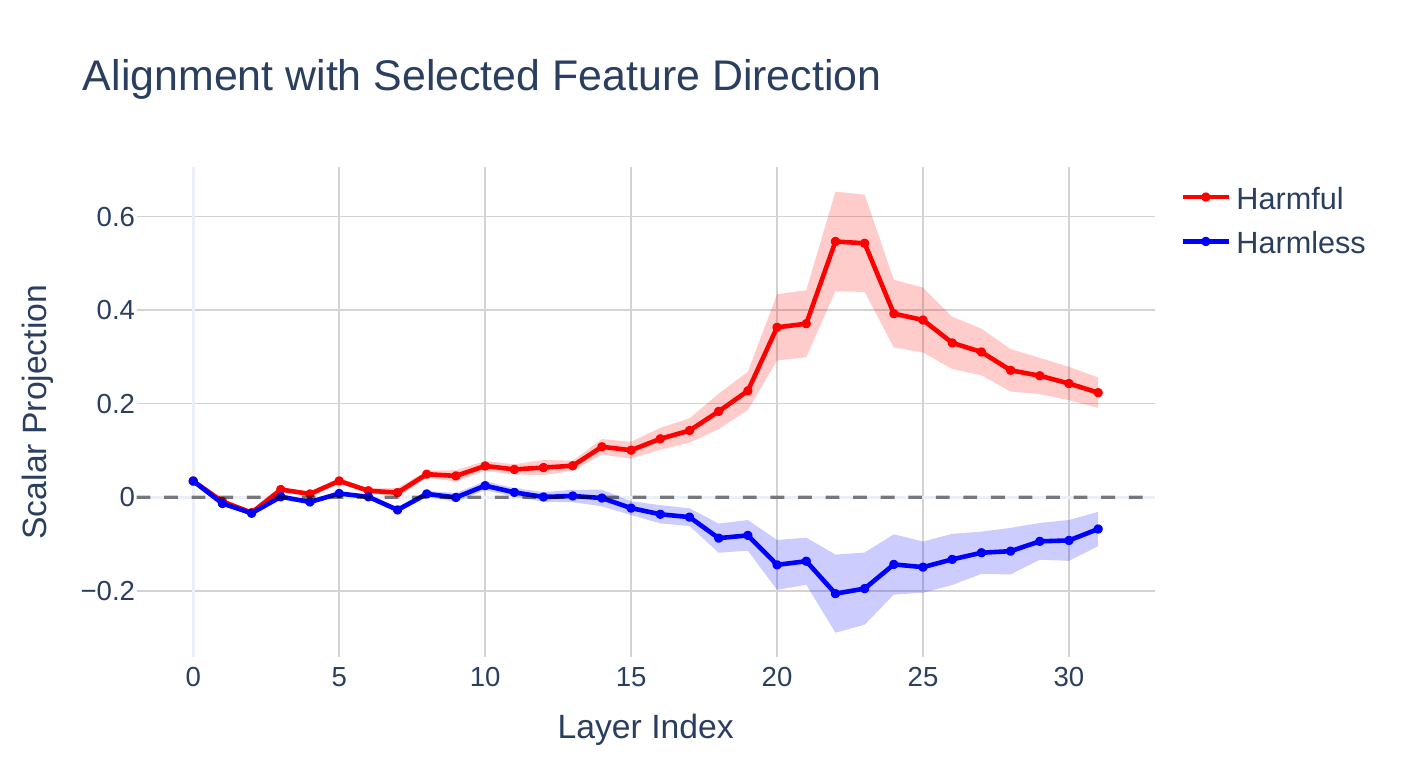}
  }
  \caption{
    \textbf{Layer-wise heterogeneity in Llama-3.2-1B-Instruct.} 
  }
  
\end{figure*}

\begin{figure*}[t]
  \centering
  \subfloat[Activation norms across layers]{
    \includegraphics[width=0.48\textwidth]{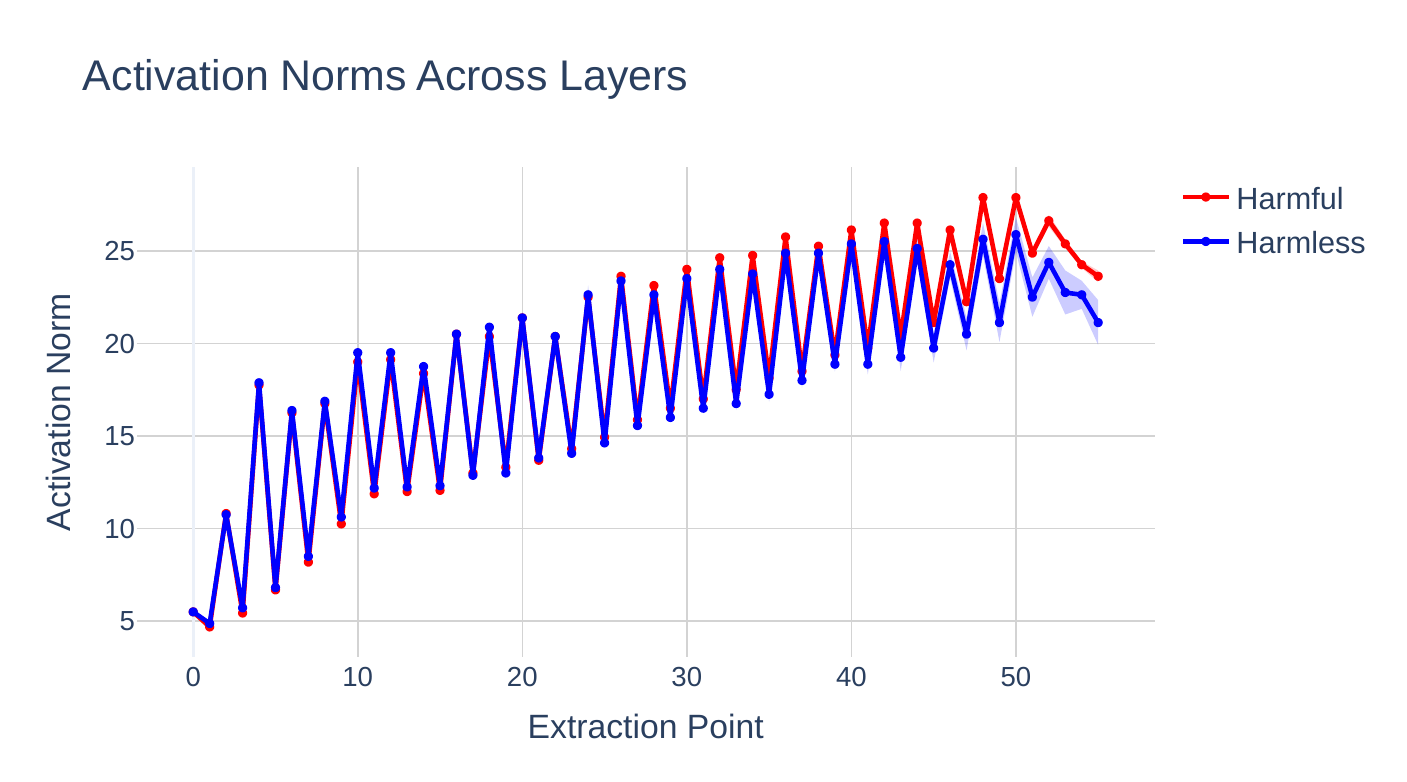}
  }\hfill
  \subfloat[Projections on local candidate directions]{
    \includegraphics[width=0.48\textwidth]{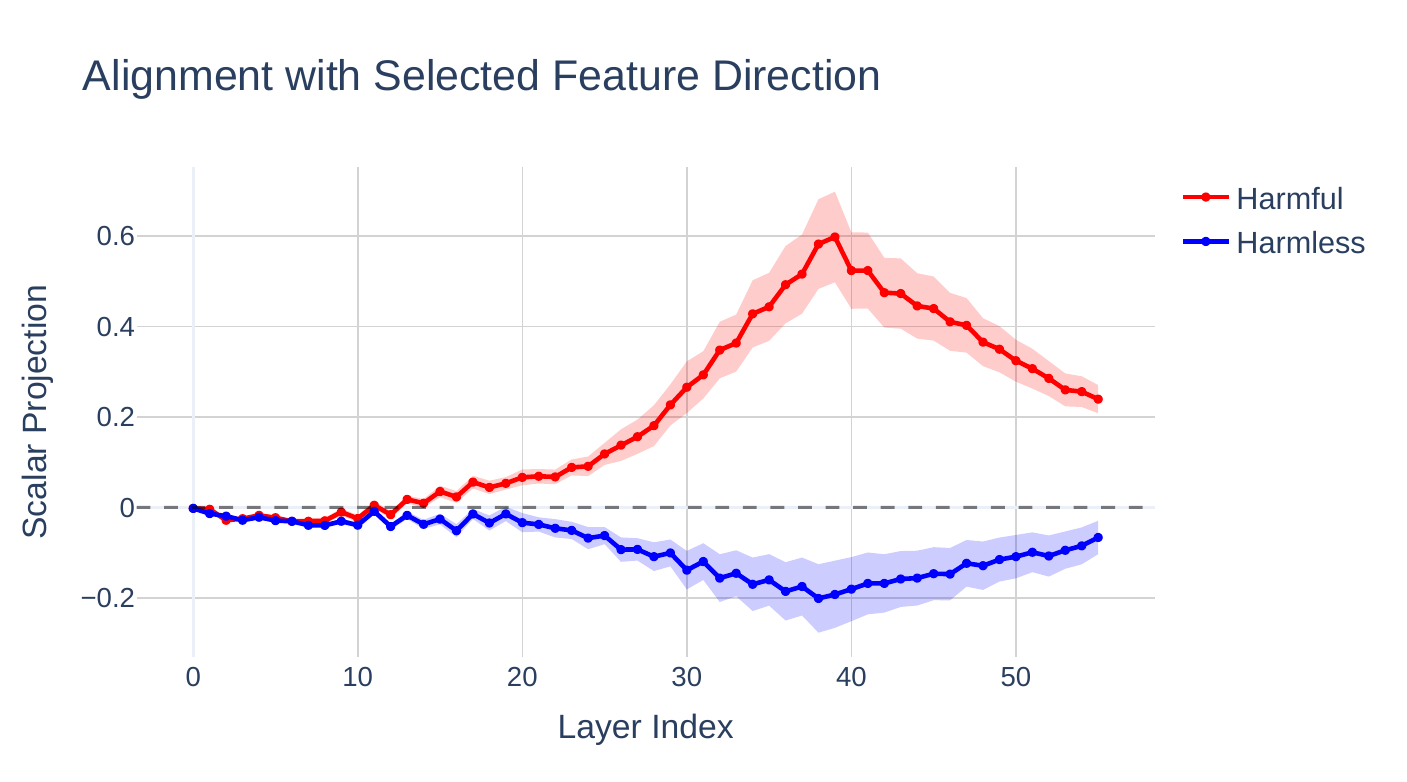}
  }
  \caption{
    \textbf{Layer-wise heterogeneity in Llama-3.2-3B-Instruct.} 
  }
  
\end{figure*}

\begin{figure*}[t]
  \centering
  \subfloat[Activation norms across layers]{
    \includegraphics[width=0.48\textwidth]{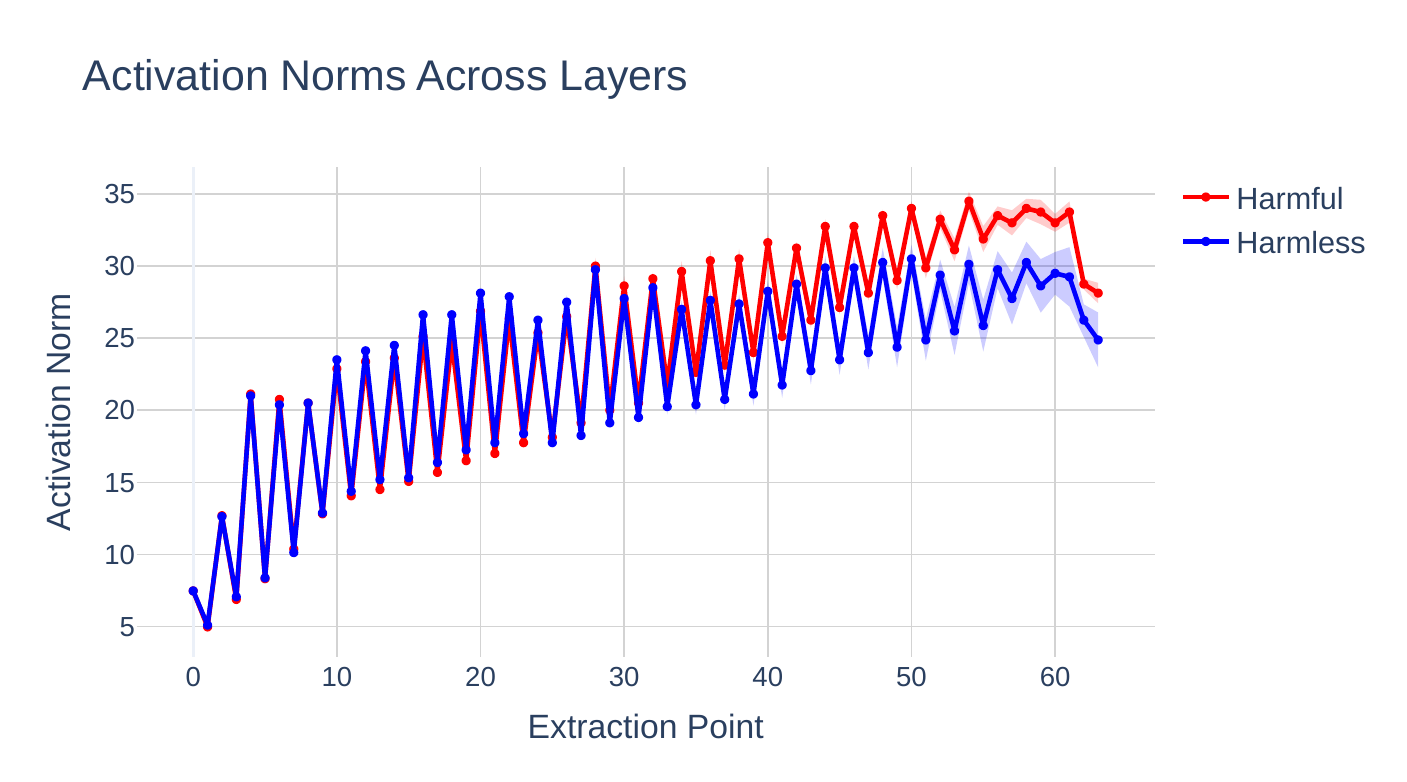}
  }\hfill
  \subfloat[Projections on local candidate directions]{
    \includegraphics[width=0.48\textwidth]{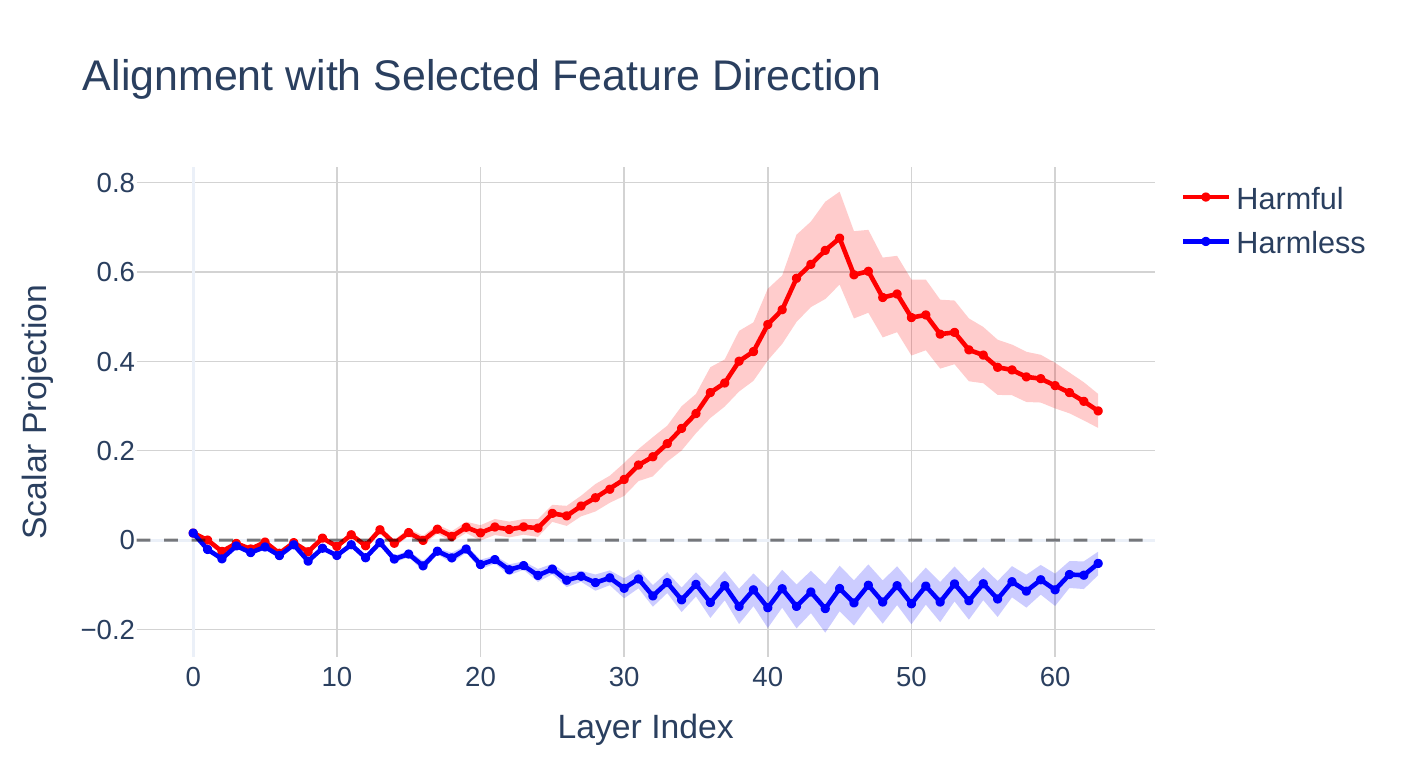}
  }
  \caption{
    \textbf{Layer-wise heterogeneity in Llama-3.1-8B-Instruct.} 
  }
  
\end{figure*}

\begin{figure*}[t]
  \centering
  \subfloat[Activation norms across layers]{
    \includegraphics[width=0.48\textwidth]{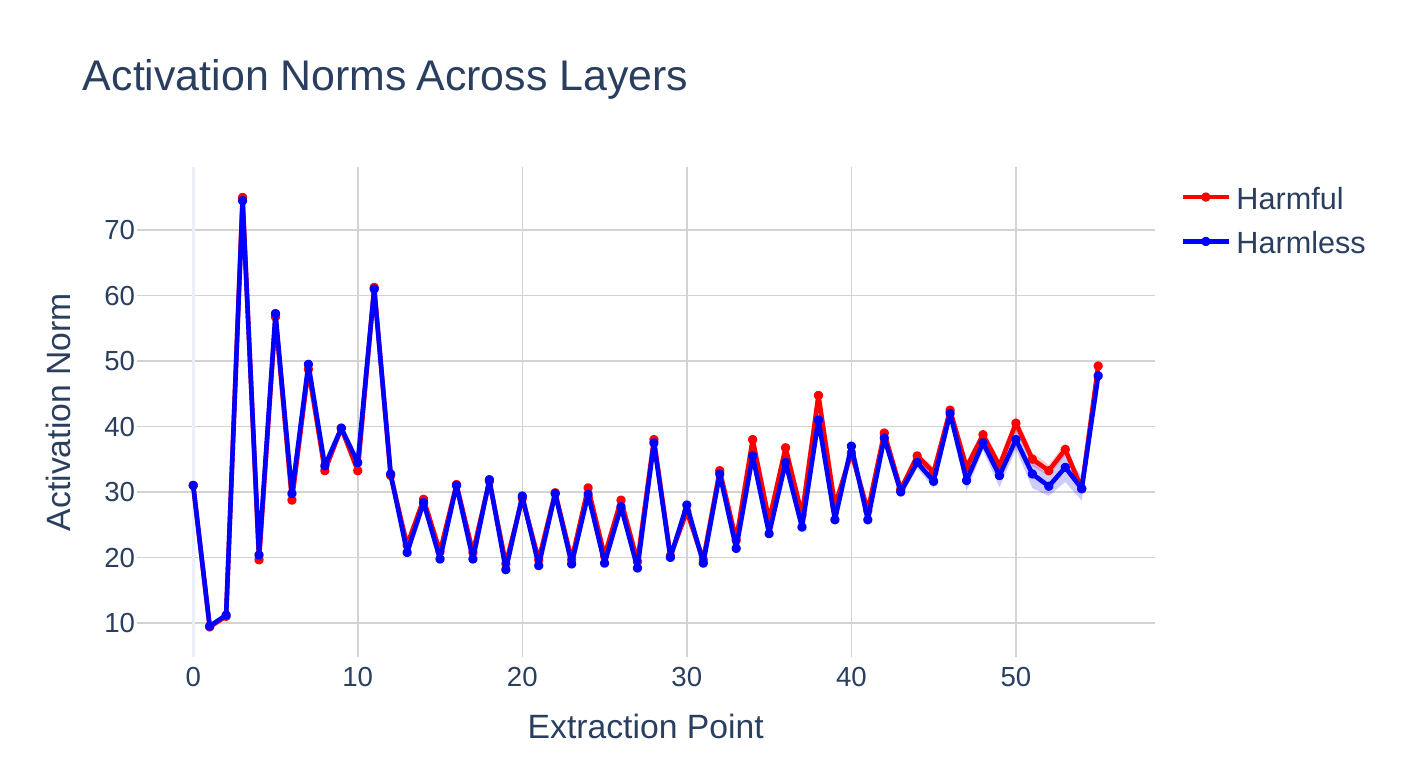}
  }\hfill
  \subfloat[Projections on local candidate directions]{
    \includegraphics[width=0.48\textwidth]{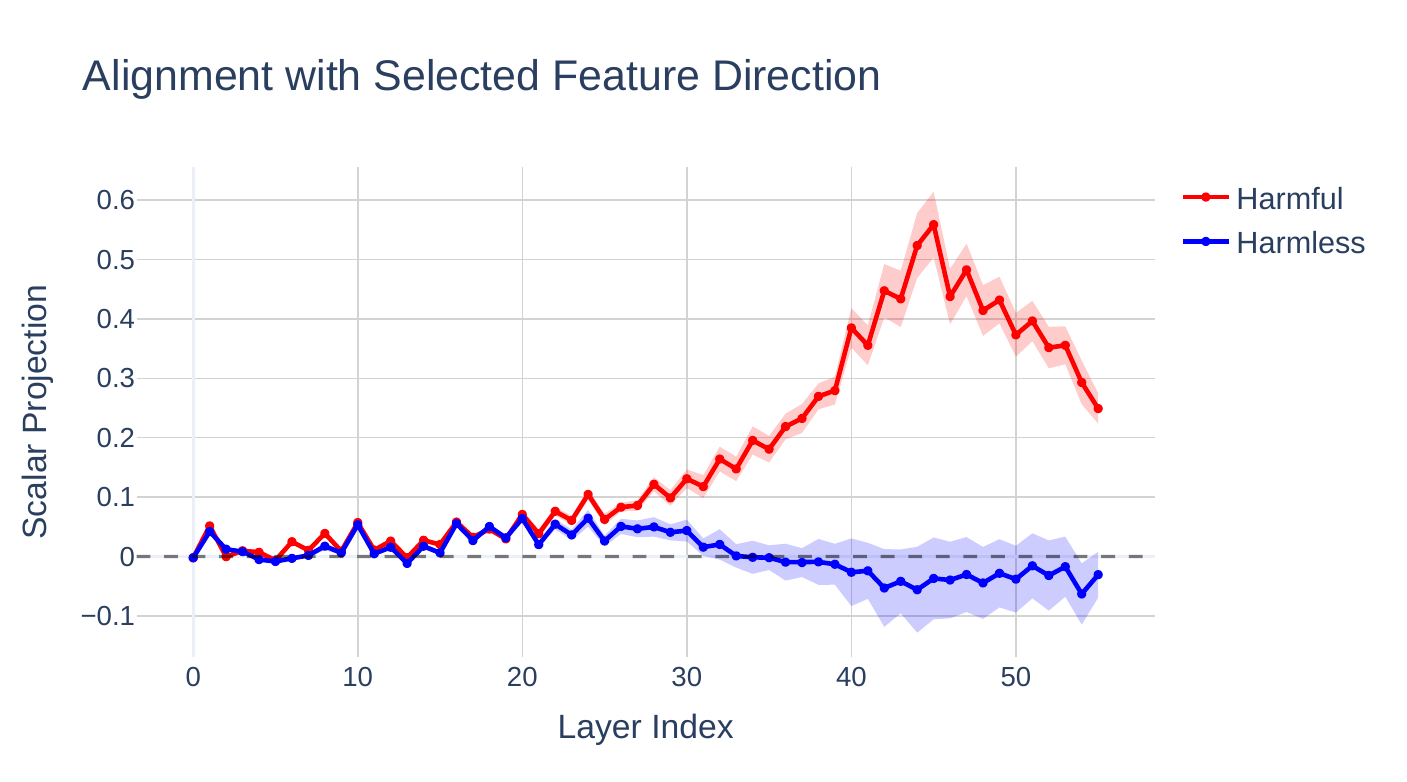}
  }
  \caption{
    \textbf{Layer-wise heterogeneity in Qwen2.5-1.5B-Instruct.} 
  }
  
\end{figure*}

\begin{figure*}[t]
  \centering
  \subfloat[Activation norms across layers]{
    \includegraphics[width=0.48\textwidth]{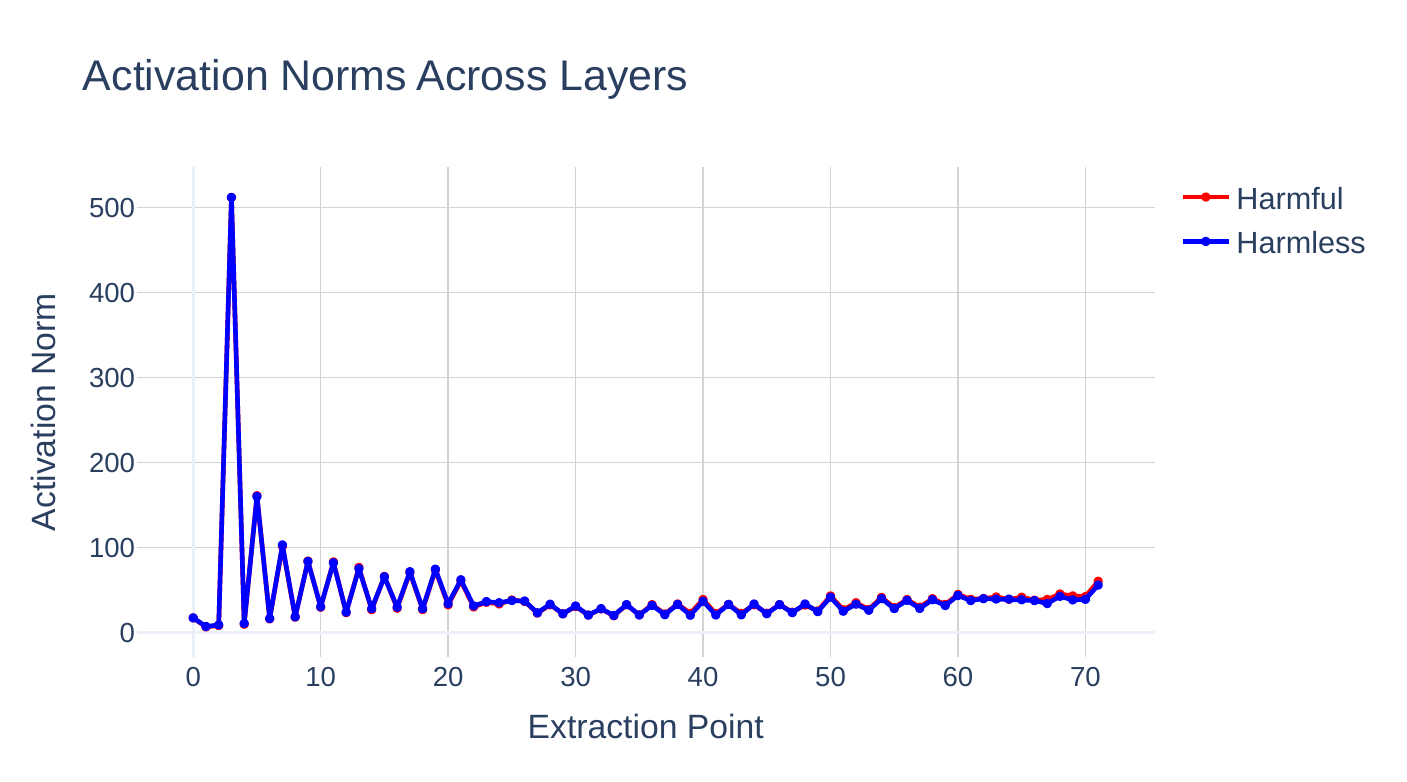}
  }\hfill
  \subfloat[Projections on local candidate directions]{
    \includegraphics[width=0.48\textwidth]{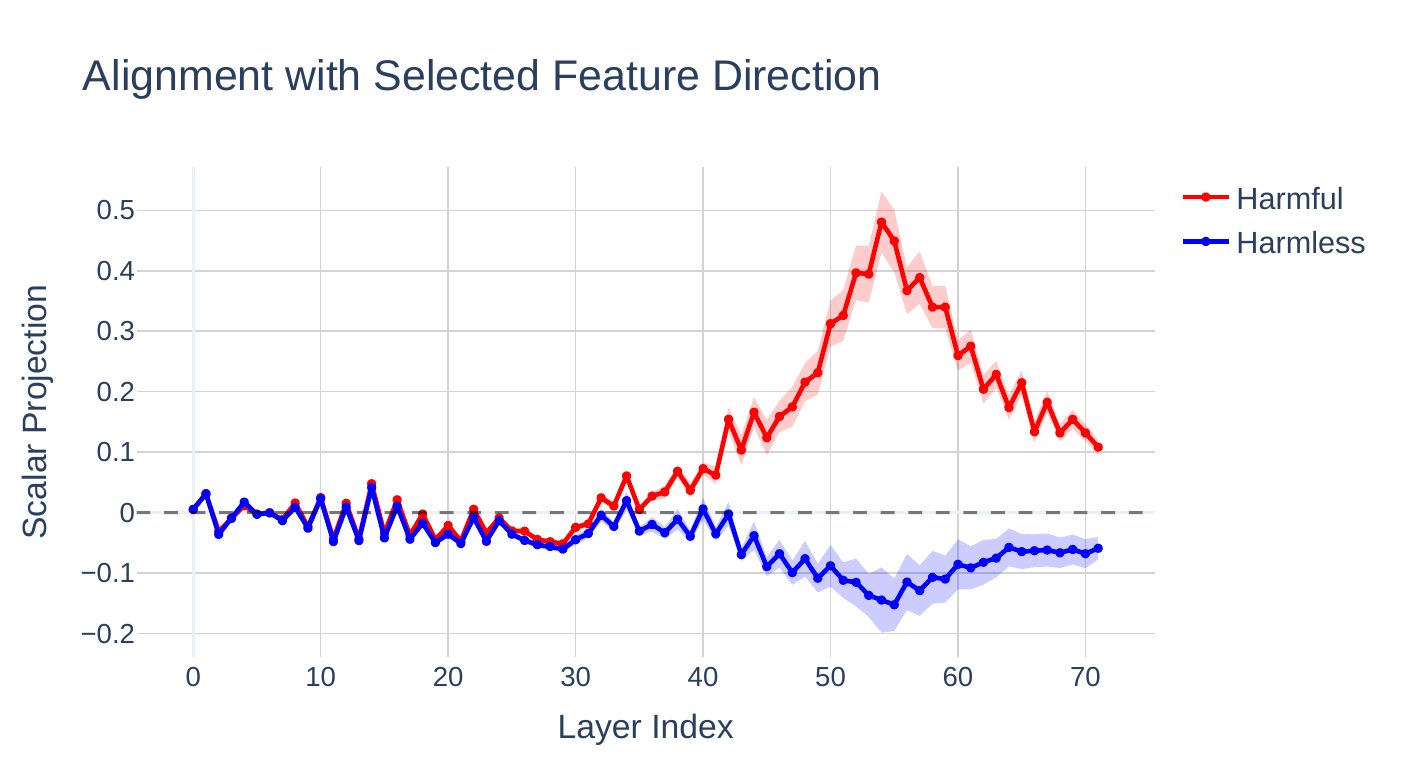}
  }
  \caption{
    \textbf{Layer-wise heterogeneity in Qwen2.5-3B-Instruct.} 
  }
\end{figure*}

\begin{figure*}[t]
  \centering
  \subfloat[Activation norms across layers]{
    \includegraphics[width=0.48\textwidth]{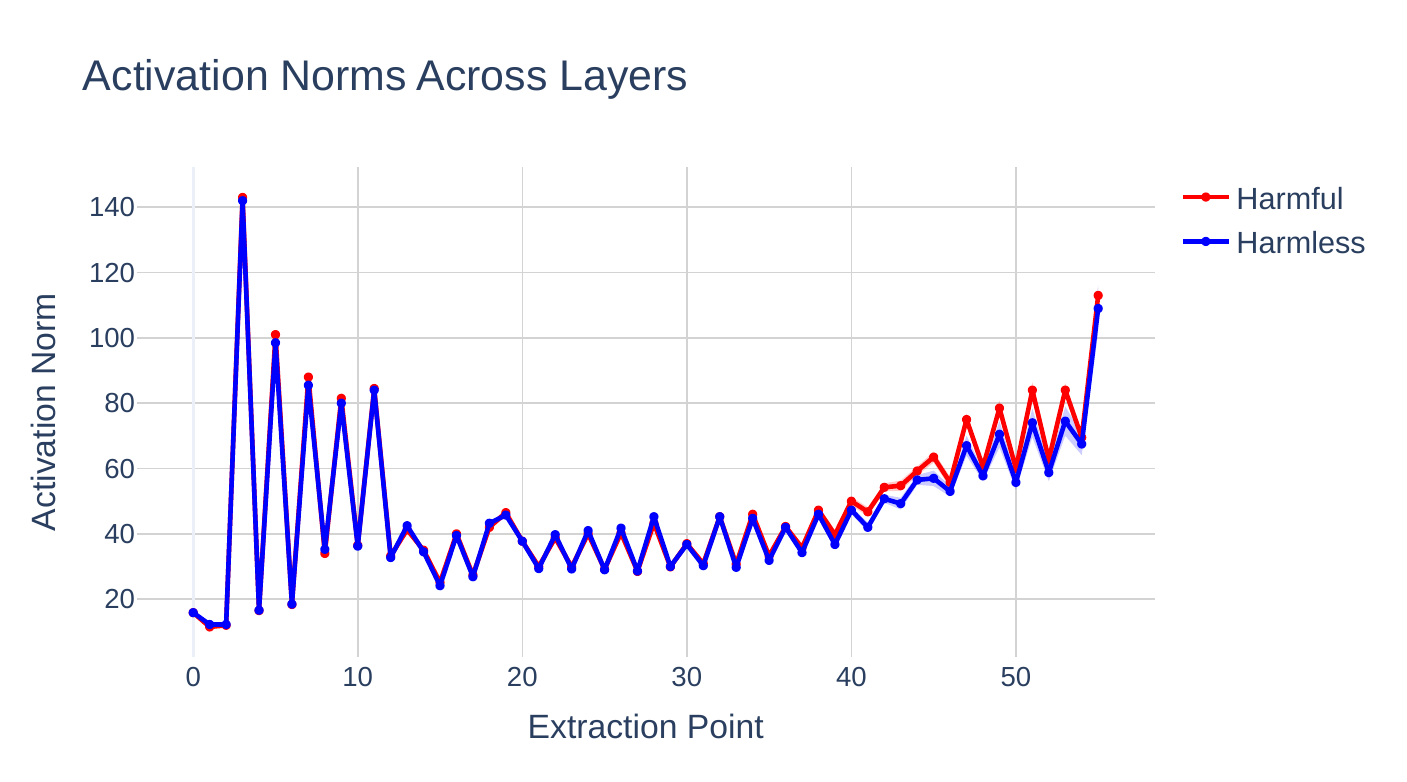}
  }\hfill
  \subfloat[Projections on local candidate directions]{
    \includegraphics[width=0.48\textwidth]{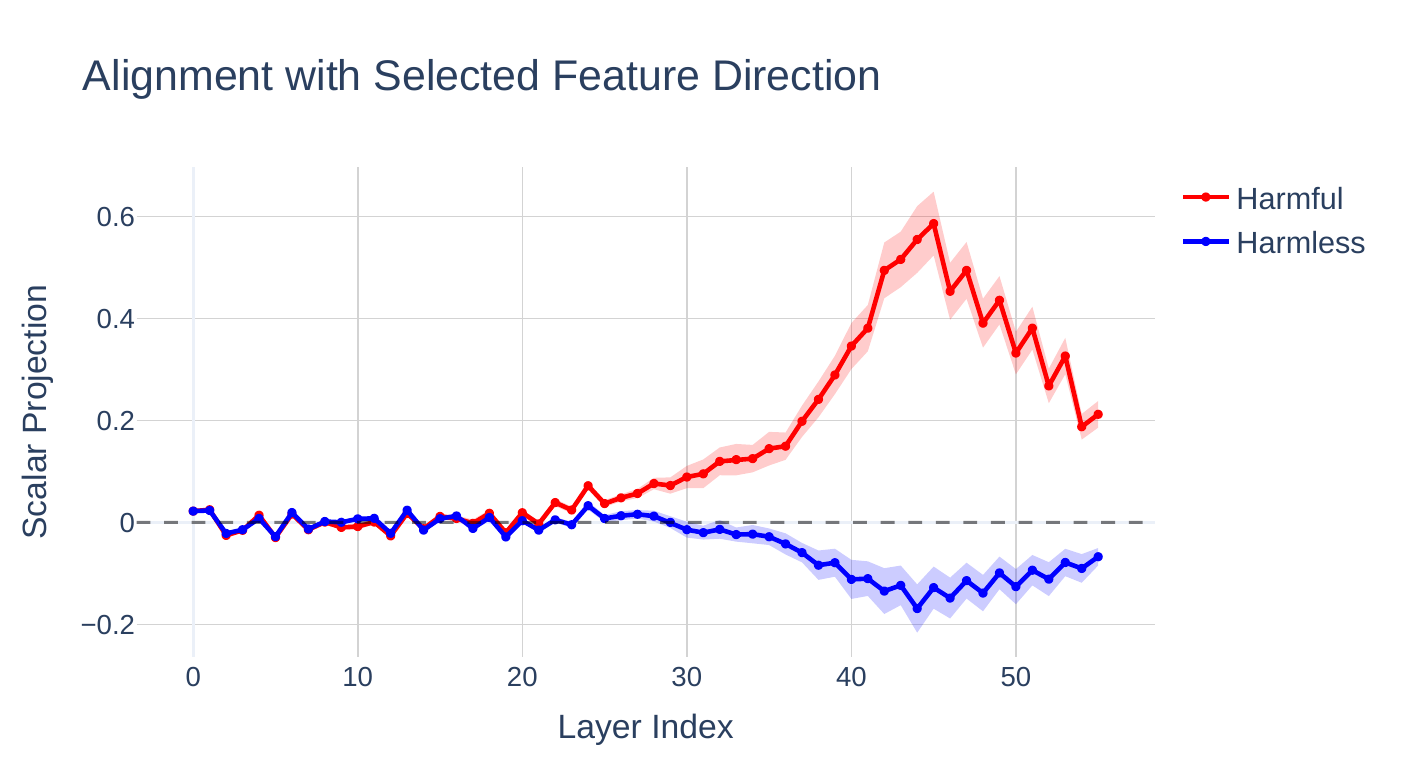}
  }
  \caption{
    \textbf{Layer-wise heterogeneity in Qwen2.5-7B-Instruct.} 
  }
\end{figure*}

\end{document}